\documentclass[12pt]{article}
\usepackage{scicite}
\usepackage{times}
\usepackage[utf8]{inputenc} 
\usepackage[T1]{fontenc}    
\usepackage{url}            
\usepackage{booktabs}       
\usepackage{nicefrac}       
\usepackage{microtype}      
\usepackage{graphicx,color}
\usepackage{float}
\graphicspath{ {./Figure/} }
\usepackage[labelfont=bf]{caption}
\DeclareGraphicsExtensions{.pdf}
\usepackage{booktabs}

\usepackage{enumitem}

\usepackage{multirow}
\usepackage{multicol}
\usepackage{amsfonts}       
\usepackage{amsmath}       
\usepackage{amssymb}

\usepackage{pdfpages}

\newcommand{\fm}[1]{{\bf #1}}

\newcommand{\FigS}[2]{Figure~\ref{#1}\,#2}  
\newcommand{\fig}[1]{Fig.~\ref{#1}}    
\newcommand{\figS}[2]{Fig.~\ref{#1}\,#2}    
\newcommand{\tab}[1]{Table~\ref{#1}}

\newcommand{\eqn}[1]{Eq.~\ref{#1}} 
\newcommand{\eqnp}[1]{(Eq.~\ref{#1})} 
\renewcommand{\sec}[1]{Sec.~\ref{#1}} 

\usepackage{xspace}
\newtheorem{definition}{Definition}

\usepackage{xr}
\makeatletter
\newcommand*{\addFileDependency}[1]{
  \typeout{(#1)}
  \@addtofilelist{#1}
  \IfFileExists{#1}{}{\typeout{No file #1.}}
}
\makeatother

\makeatletter
\DeclareRobustCommand\onedot{\futurelet\@let@token\@onedot}
\def\@onedot{\ifx\@let@token.\else.\null\fi\xspace}
\def\eg{e.g\onedot}
\def\ie{i.e\onedot}
\def\cf{cf\onedot}

\def\wrt{w.r.t\onedot}
\makeatother

\usepackage{lineno}

\usepackage{xcolor}
\definecolor{ourblue}{rgb}{0.368,0.507,0.71}
\definecolor{ourorange}{rgb}{0.881,0.611,0.142}
\definecolor{ourgreen}{rgb}{0.56,0.692,0.195}
\definecolor{ourred}{rgb}{0.923,0.386,0.209}
\definecolor{ourviolet}{rgb}{0.528,0.471,0.701}
\definecolor{ourbrown}{rgb}{0.772,0.432,0.102}
\definecolor{ourlightblue}{rgb}{0.364,0.619,0.782}
\definecolor{ourdarkgreen}{rgb}{0.572,0.586,0.}

\usepackage[colorlinks=true, linkcolor=ourblue, citecolor=ourgreen,urlcolor=ourblue]{hyperref}

\newenvironment{sciabstract}{%
\begin{quote} \bf}
{\end{quote}}

\title{Theory and Design of Super-resolution Haptic Skins} 

\author{Huanbo Sun$^{1\ast}$ and Georg Martius$^{1\ast}$\\
        \normalsize{$^{1}$ Autonomous Learning Group,}\\[-.2em]
        \normalsize{Max Planck Institute for Intelligent Systems, T\"ubingen, Germany.}\\
        \normalsize{$^\ast$ Corresponding author. Emails: huanbo.sun | georg.martius@tuebingen.mpg.de}}

\begin{document} 
\maketitle 
%

\begin{sciabstract} 
Haptic feedback is important to make robots more dexterous and effective in unstructured environments.
High-resolution haptic sensors are still not widely available, and their application is often bound by the resolution-robustness dilemma.
A route towards high-resolution and robust skin embeds a few sensor units (taxels) into a flexible surface material and uses signal processing to achieve sensing with super-resolution accuracy.
We propose a theory for geometric super-resolution to guide the development of haptic sensors of this kind and link it to machine learning techniques for signal processing.
This theory is based on sensor isolines and allows us to predict force sensitivity and accuracy in contact position and force magnitude as a spatial quantity.
We evaluate the influence of different factors, such as elastic properties of the material, structure design, and transduction methods, using finite element simulations and by implementing real sensors.
We empirically determine sensor isolines and validate the theory in two custom-built sensors with barometric units for 1D and 2D measurement surfaces.
Using machine learning methods for the inference of contact information, our sensors obtain an unparalleled average super-resolution factor of over 100 and 1200, respectively.
Our theory can guide future haptic sensor designs and inform various design choices.
\end{sciabstract}

\pagebreak
\section{Introduction}
Haptic sensors are indispensable in robotic applications to enable robots to perceive when, where, and how their bodies are interactively contacting other things~\cite{Review1,Review2,Review3,Review4,Review5,Review6}.
A common theme for surface haptic sensors is to integrate many small sensing units forming a type of grid along a flat or curved surface.
Each sensing unit, named taxel, is responsible for sensing interactions near its location.
For typical applications, a resolution is desirable that would imply numerous taxels~\cite{Review4,Review5,Art1,Art2,Art3,Art4,Art5}.
This is true both for small surface sensing, \eg, at the fingertips, and for large surfaces, \eg, around limbs.
For fingertips, very focused areas need a high density of taxels with a fine size to perceive high-resolution haptic information, which is similar to touch screens.
For limbs, even though the haptic information is coarsely needed referring to the density distribution of human mechanoreceptors~\cite{Mechanoreceptors}, but the sensing areas are comparably large which also need many taxels.
Technical challenges arise concerning the physical size of the taxels as well as growing manufacturing and wiring costs~\cite{Requirements}.

Naturally, efforts have been made to reduce the number of taxels~\cite{BioTac,GSR,SG1,SG2,ERT1,ERT2,SR_Magnet}.
The reduction is generally enabled by the fact that a taxel can monitor not only a tiny area but an extended patch on the surface and that we are interested in a subset of all possible stimulation patterns, for instance, a few touch-points.
The particular material properties lead to a characteristic spread of contact information to the sensing taxels. 
Different physical effects can be used, such as geometric and mechanical properties~\cite{GSR,SG1,SG2}, electrical resistance~\cite{ERT1,ERT2}, magnetic flux~\cite{SR_Magnet}, thermo- and fluid dynamics~\cite{MIGP}, and so forth.
The central idea is to solve the inverse problem of inferring haptic information from a few sensors, effectively creating high-resolution virtual taxels. 
This is also referred to as super-resolution sensing~\cite{VirtualSensing}.

Super-resolution sensing was early explored in different areas, such as geostatistics (temperature and precipitation prediction)~\cite{MIGP,Climate}, imaging systems (super-resolution microscopy)~\cite{OSR,Microscopy}, and material releases detection (aerosol/chemical plume release)~\cite{MaterialRelease,ChemicalRelease}.
They typically rely on the continuity and neighboring effects in the transmission medium and on solving inverse problems (signal processing) to reconstruct spatial information from a few sensors in a sparse configuration.
There are also several haptic sensor designs that achieve super-resolution sensing. 
HapDef~\cite{SG2} makes use of the spreading behavior of mechanical deformation by attaching a few strain gauges on a large robotic limb shell and developing machine learning (ML) methods to achieve a 78-fold super-resolution (SR).
Similarly, Piacenza et al.~\cite{Baro} mold a few barometers inside silicone and employs ML to achieve a 57-fold SR.
Lepora et al.~\cite{GSR} customizes a sensor with several capacitive taxels and applies Bayesian optimization to achieve a 35-fold SR.
Yan et al.~\cite{SR_Magnet} uses Hall sensors in sparse configuration to detect the deformation of a magnetized flexible film and develops simple ML models to achieve a 30-fold SR.
These sensors rely on the empirical knowledge that overlapping multiple taxels' perception fields triggers SR behavior.
However, a sounded quantitative theory is still missing in solving the inverse problem and providing a basis for a fast back-projection-like approach towards system-level sensor designs.
The theory would imply a baseline for evaluating signal-processing methods from a model-centric perspective (optimizing ML models based on whatever data collected), and offer guidelines of data collection for data-driven sensors from a data-centric perspective (choosing which data to collect given ML models).

\begin{figure}
    \centering
    \includegraphics[width=\textwidth]{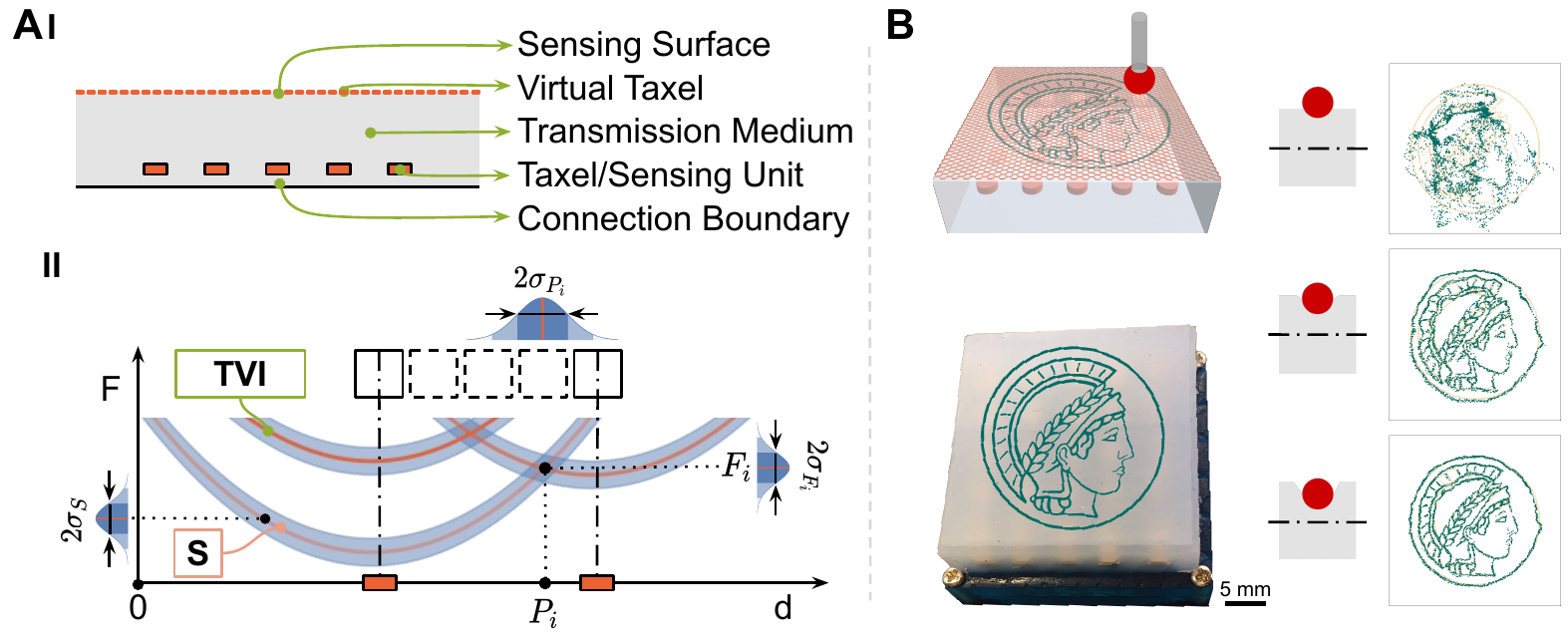}
    \caption{Introduction of the model.
    \fm{A-I} depicts a 1D sensor layout with sparsely placed real taxels and highly resolved virtual taxels.
    \fm{A-II} shows the orange taxel value isolines (TVIs) that are the force $F$ needed to elicit the same taxel-value $S$ depending on the distance $d$ from the sensor unit.
    The intersection of TVIs of two sensors indicates a hypothetical contact point $P_i$ with a force strength $F_i$.
    The measurement noise $\sigma_S$ in shaded blue leads to uncertainty of the position $\sigma_{P_i}$ and force $\sigma_{F_i}$.
    \fm{B} shows a validation of our theory on a 2D sensor that embeds 25 barometers in an elastomer (EcoFlex 00-30) with a grid layout (5 $\times$ 5) to realize the super-resolution functionality.
    As a demonstration, a 4\,mm red spherical indenter contacts the square sensor surface (34\,mm $\times$ 34\,mm) following the ``Minerva'' pattern (diameter 26\,mm) with different depths (1.2\,mm, 2.4\,mm, 3.6\,mm). The sensor can resolve the contact pattern in different resolution depending on indentation depths.
    }\label{fig:Introduction}
\end{figure}

How much ``super-resolution'' can we obtain?
What is the expected sensitivity of the sensing system?
Which surface material and which transduction method is suitable?
Why and how can machine learning help the inference?
This paper aims to provide answers to these questions from a theoretical perspective as well as from a practical point of view by applying the method to three common sensor types and evaluating two custom-built sensor designs.
We introduce a theory that allows us to infer spatially resolved properties, such as accuracy and sensitivity for single and double contact, based on a single characteristic of the material-taxel interplay: \textbf{taxel~value~isolines (TVIs)}.

The model we use is a sensor comprising discrete taxels sparsely distributed in a continuous transmission medium underneath a sensing surface as shown in~\fig{fig:Introduction}.
For sensing devices that can be described by this model, we derive the single-contact accuracy for position and force magnitude inference based on the taxel value isolines; determine the minimal force profile for which localization is possible; provide conditions for simultaneous contacts to be distinguishable; evaluate the influence of different factors, such as elastic properties of the material, structure design, and transduction methods as well as contacting object shapes; and design a one-dimensional (1D) and a two-dimensional (2D) barometer-based sensors to validate our proposed super-resolution theory.

\section{Results}
\subsection{The Model}
We consider a class of haptic sensing devices intended for measuring force interactions on an extended surface that have an elastic transmission medium covering or embedding physical sensor units (taxels).
A one-dimensional model is shown in \figS{fig:Introduction}{A-I}.
A single taxel value $s$ is a function of the applied contact force strength $F$, and the displacement $d$ between the contact center and the taxel center:
\begin{equation}
    s = f(F,d) + \epsilon_S,\label{eqn:taxelvalue}
\end{equation}
where $\epsilon_S$ is the taxel measurement noise with a constant standard deviation $\sigma_S$:
\begin{equation}
    \epsilon_S \sim \mathcal{N}(0, \sigma_S^2).\label{eqn:taxelvalue-noise}
\end{equation}
To relate the sensor measurement noise $\epsilon_S$ to force values $F$, we assume a linear relationship.
Formally, the sensor responses to an applied force at distance 0 should be
$f(F,0) = c\cdot F$ and to simplify the notation we set $c=1$.
However, our analysis can be easily adapted for different values of $c$ and for nonlinear monotonous relationships. 
A detailed analysis of $c$ for a sensor type used in this paper is provided in \sec{sec:sup:sensorforcerelationship}, which supports our linearity assumption.
One of our main contributions is to introduce the \emph{taxel value isolines} as an important characteristic function of the system.

\begin{definition}{\it 
\emph{Taxel value isolines (TVIs)} are a family of curves }
\begin{align}
I^S(d) = \begin{cases} F &\text{with } f(F, d) = S \\ \text{undefined} &\text{if no such } F \end{cases}
\end{align}{\it
where the mean taxel output \eqnp{eqn:taxelvalue} has a constant value $S$.}
\end{definition}

The TVI for the model system and the effect of measurement noise are shown in \figS{fig:Introduction}{A-II}.
Intuitively, the isolines quantify how much force is needed along the surface to yield the same sensor value. To activate a taxel with a particular value, the required force strength is smaller when the contact location is closer to the taxel.
Based on these isolines, we can derive the accuracy distributions of contact localization, force quantification, and the distribution of sensitivity over the sensing surface.
Our method is based on the conditions for unique nonlinear triangulation and the geometry of intersecting isolines with their uncertainty bands.

\subsection{Super-resolution in 1D}
\begin{figure}
    \centering
    \vspace{-2cm}
    \includegraphics[width=\columnwidth]{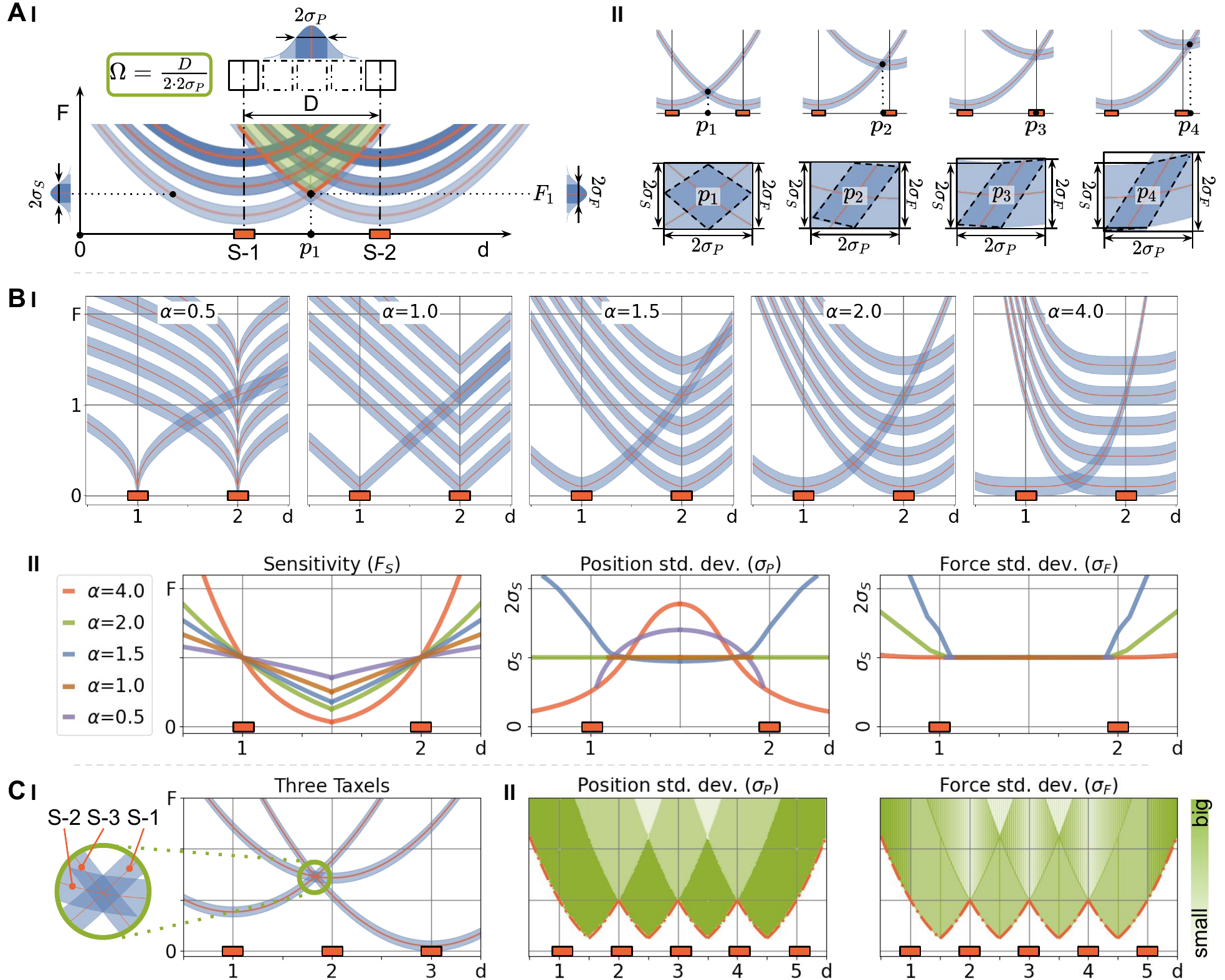}\\[1em]
    \caption{Theory of super-resolution in 1D. 
    \fm{A}: Model for contact point localization at super-resolution ($\Omega$).
    \fm{A-I} shows the intersection of TVIs of two sensors (marked as S-1 and S-2).
    The place where the lines corresponding to the particular sensor readings cross is the contact location that is marked as $p_1$.
    The measurement noise $\sigma_S$ leads to uncertainty of position $\sigma_P$ and force $\sigma_F$.
    \fm{A-II}: Different intersection types.
    \fm{B}: Effect of TVIs shape on super-resolution characteristics.
    \fm{B-I}: TVIs for two taxels (at distance 1) with different attenuation exponents $\alpha$.
    Note the different shapes of the intersection areas.
    \fm{B-II}: Resulting sensitivity, standard deviation of position localization and standard deviation of force inference.
    \fm{C}: Theoretical super-resolution characteristics of a 1D sensor with multiple taxels.
    \fm{C-I}: Three taxels localizing a single contact.
    \fm{C-II}: The spatial distribution of accuracy for a single contact.
    Below the orange dash-dotted line, no super-resolution localization is possible.
    Notice the increase in accuracy for higher forces because multiple taxels are activated.
    }\label{fig:Theory1D}
\end{figure}

When can a single contact point be localized at super-resolution, meaning much more accurate than the distance between taxels?
Intuitively, this is possible when two or more taxels measure nonzero responses to the contact force and the activation pattern of the taxels is unique for this location.
More formally, this condition can be analyzed with the TVIs. 
The minimal setting of two taxels located at a distance of $D$ is shown in \fig{fig:Theory1D}.

A particular contact event causes a sensor reading in both sensors S-1 and S-2. 
The TVI corresponding to a sensor reading of S-1 relates the position to the force of the potential contact point.
Only if the TVIs from both sensors intersect, the contact position can be localized, up to some uncertainty introduced by the measurement noise $\sigma_S$ \eqnp{eqn:taxelvalue}.
With this insight, we can derive the minimal force sensitivity $F_S$ that allows for super-resolution localization, \ie where at least two isolines intersect.
We can find this minimal force, where the isolines corresponding to the smallest taxel sensitivity $S_\text{min}$ intersects with TVIs from neighboring taxels.
The area where $F\geq F_S$ is shaded in green in \figS{fig:Theory1D}{A-I}.

As we assume an additive measurement noise, the uncertainty is also added to the isolines.
What does this mean for the ability to localize the contact position and to infer force magnitude?
For the case of parabolic TVIs, we have to consider different possible scenarios, as shown in \figS{fig:Theory1D}{A-II}. 
For contact locations between the two taxels, marked with $p_1, p_2$, the standard deviation for the force estimation is actually constant and given by $\sigma_S$.
There is no dependency on the taxel distance $D$.
For contact points close to the taxel and outside the taxel pair, the uncertainty of the force magnitude grows with distance ($p_3, p_4$). 
The standard deviation for position localization $\sigma_P$ is typically largest in the center between the taxels and gets smaller on either side (exceptions are discussed in the following).
The exact equations to compute these standard deviations are given in~\sec{sec:sup:sigmas} and visualized~\figS{fig:Theory1D}{B}, as described below.

Knowing the position accuracy $\sigma_P$, we can quantify the super-resolution capabilities. 
For simplicity, we use $2\sigma_P$ as the size of a virtual taxel, as shown in \figS{fig:Theory1D}{A-I}, corresponding to a confidence interval of about $68\%$.
Thus, the spatial resolution is $2\sigma_P$ and between two sensors at a distance $D$ we can distinguish $D/2\sigma_P$ virtual taxels, which we define as \emph{super-resolution factor} \wrt real taxel number $n$ ($n=2$ in \figS{fig:Theory1D}{A-I}):

\begin{equation}
    \Omega = \frac{D}{n \cdot 2\sigma_P}. \label{eqn:superresfactor}
\end{equation}

\paragraph{Influence of the isoline shape}\label{sec:1D:convexity}
The shape of the TVIs depends on the properties of the transmission medium and the sensor unit type. 
We study the impact of the TVI shape on the accuracy of single contact force inference.
We follow the general model that the response of a taxel to a force on the surface decreases monotonously with distance from the taxel center.
This in turn leads to a monotonously increasing TVI.

We assume the attenuation of taxel response behaves as $s \propto 1/|d|^\alpha$ for a force at distance $|d|$ from the taxel center with the attenuation exponent $\alpha$. 
Without loss of generality, we consider two taxels at a distance of 1. 
Their TVIs are given by
\begin{align}
        I^{S_1}_1(d) &= g(S_1) +  |d|^{\alpha},\\
        I^{S_2}_2(d,s) &= g(S_2) + |1-d|^{\alpha}.
\end{align}
where $g(S)$ is the force corresponding to the measurement at distance 0.
The TVIs for different attenuation behaviors are shown in \figS{fig:Theory1D}{B}.
The accuracy of super-resolution sensing is strongly affected by $\alpha$.
For instance, for linear attenuation ($\alpha=1$), the localization will only be possible between the taxels, but not outside (unbounded intersection area).
For concave curves ($\alpha<1$), there can be three intersection areas: one is between the two taxels and the other two are outside, which makes the reconstruction not unique.
For the sake of comparison, we assume that in this case the correct intersection is known. 
 
Smaller attenuation exponent $\alpha$ yields a smaller sensitivity (bigger $F_S$) between the taxels but in general a more homogeneous distribution (\figS{fig:Theory1D}{B-II}).
The position accuracy, measured by the standard deviation, $\sigma_P$ is constant for $\alpha=2$. 
For normalized TVIs, as used here for comparison, $\sigma_P = \sigma_S$.
The general case is given below. 
For $\alpha>2$ and $\alpha<1$ we find a reduced position accuracy between the taxels. 
Interestingly, for $\alpha>2$, the reduced position error at the outside is paid by a reduced force sensitivity.
The ability to infer the force magnitude, given by $\sigma_F$, is constant between the two taxels, irrespective of $\alpha$.
For contact points outside, larger values of $\alpha$ are better, whereas for $\alpha \leq 1$ no detection is possible.
For $1<\alpha<2$, the position error and force error are both high outside.
From this analysis, $\alpha=2$ yields the best overall characteristics, namely having high force sensitivity (small $F_S$), high position, and force accuracy (small $\sigma_P, \sigma_F$ respectively). 
Values $\alpha>2$ trade off some super-resolution by sensitivity, but are generally also good. 
Taxel response attenuation with exponent $\alpha<2$ should be avoided, if possible.

An approximate relationship between $\sigma_P$ and $\sigma_S$ for the more general TVIs
$I^S(d) = g(S) +  \lambda |d|^{\alpha}$ is given by:
\begin{align}
    \sigma_P &= \frac{2\sigma_S}{m_1 + m_2}
\end{align}
where $m_1=\lambda \alpha |d|^{\alpha -1 }$ and $m_2=\lambda \alpha |D-d|^{\alpha -1 }$ are the absolute derivatives of the TVIs as detailed in \sec{sec:sup:approxsigmaP}. 
Using the position accuracy at $d=\nicefrac D 2$ (which is the worst for typical exponents, $\alpha \ge 2)$ for $n=2$ sensors we can compute the super-resolution factor analytically using \eqn{eqn:superresfactor}:
\begin{align}
    \Omega &= \frac{D \lambda \alpha (\nicefrac D 2)^{\alpha -1 }}{2 \cdot 2 \sigma_S}\label{eqn:superresfactordetail}
\end{align}

\paragraph{Multiple taxels in a line}\label{sec:1D:multi}
A real sensor should clearly contain more than 2 taxels, so let us consider multiple taxels (with quadratic isolines) equidistantly placed in the transmission medium (as shown in \figS{fig:Introduction}{A-I}).
As expected, in the area where more than two taxels respond to the contact stimulation, higher localization accuracy is possible, as shown in \figS{fig:Theory1D}{C-I}.
This leads to a reduction in uncertainty about the contact force.
The intersection area is smaller and, due to averaging independent noise measurements, the variance also reduces with an increased number of TVIs intersecting.
The sensitivity $F_S$ is shown in \figS{fig:Theory1D}{C-II} as a dash-dotted lower bound, together with the accuracy of localization $\sigma_P$ and force quantification $\sigma_F$. 
The sensitivity is not homogeneous and is higher between taxels, which might be a surprising result at first glance.
In summary, the most important take-home message is to have multiple taxels responding to a contact force because it improves accuracy.
Closer placement of taxels increases both sensitivity and accuracy, whereas reducing taxel measurement noise improves accuracy but not sensitivity.

\paragraph{Multiple contact points}
In many applications, we are interested in detecting multiple simultaneous contact points.
To detect two contact points, we need at first glance two pairs of taxels.
However, when the contact points are too close, spurious contact points would be detected due to additional intersections of TVIs.
The basic condition for successfully distinguishing them is shown in \fig{fig:sup:Multicontact}.
There should be at least two taxels between these two contacts, and at most one taxel of them that is evoked by one contact not the other.
With higher forces applied by these two contacts, the distinguishable distance for double simultaneous contacts is larger.

\subsection{Physical Implementation in 1D}
External contact at the sensing surface causes deformation of the transmission medium (TM) that can be measured by physical sensor units (taxels).
On the one hand, we investigate the influences of taxel placement and TM (e.g., elastomer) properties on the TVIs and on the sensitivity.
On the other hand, we augment our theoretical analysis by measuring the response curves of three suitable real physical sensors and choose one of them to validate our proposed theory.

\paragraph{Physical factors of TM influencing TVIs}
Where to place the sensor units within the transmission medium?
How thick should be the transmission medium?
What is the effect of material properties such as the Poisson's ratio?
A static mechanical model simulated with the finite element method (FEM) using Ansys~\cite{ANSYS} is built to answer the above questions.
Details are explained in~\sec{sec:sup:fem}.
It indicates that the dependency of the deformation on the depth has an impact on the TVIs, and suggests placing sensors more close to the sensing surface for higher sensitivity.
Considering the structure design, thicker material causes more displacement and in turn increases sensitivity, because the fixed boundary effect becomes less dominant.
From the material property perspective, the Young's modulus and the Poisson's ratio are the two main properties of the TM we consider.
The Young's modulus describes how easy the material deforms and has a proportional impact on the deformation.
A soft material (small Young's modulus) improves sensitivity but is also increasingly deformed by inertial effects, which needs a trade-off with the material density.
The Poisson's ratio measures the relative transverse/axial expansion when the material is axially compressed.
Most elastomers have Poisson's ratio around 0.5, and metals have around 0.3~\cite{PoissonRatio}.
Decreasing the ratio, the radial displacement becomes much less sensitive, whereas the depth displacement has higher sensitivity (lower TVIs).
Thus, depending on the measurement direction of the real sensor unit, different Poisson's ratios are preferred.

\begin{figure}[thp]
    \centering
    \includegraphics[width=\columnwidth]{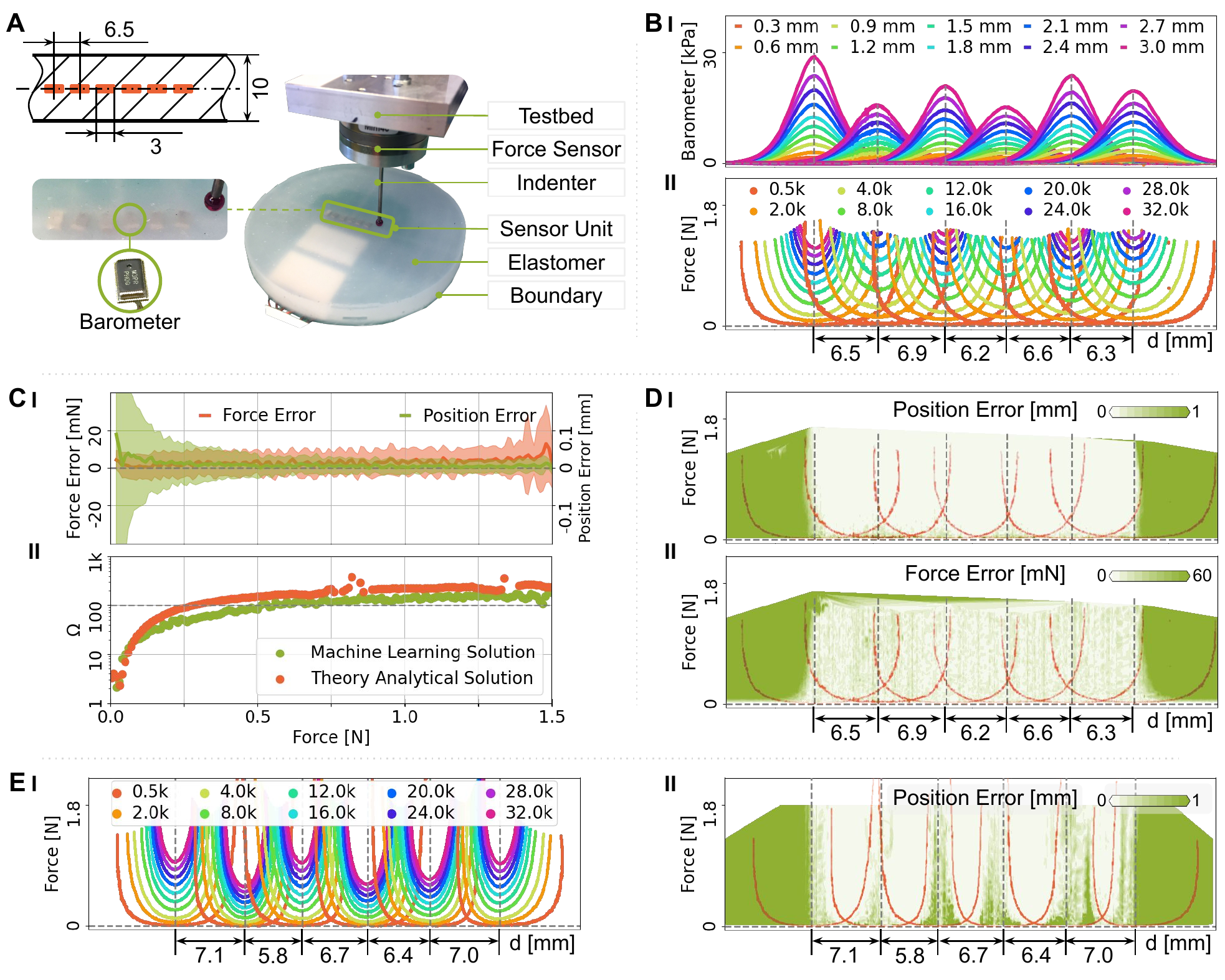}\\[-.8em]
    \caption{Super-resolution in 1D. 
    Real line-sensor device with six barometer taxels at a distance of about 6.5\,mm.
        \fm{A}: Sensor device with stimulation testbed.
        \fm{B}: Taxel response for different indentation depths (\fm{I}) and resulting TVIs for different sensor values (\fm{II}).
        \fm{C, D}: Quantitative evaluation of the sensor device using machine learning models to infer the position and force magnitude.
        \fm{C-I}: Position and force error depending on stimulation force: mean and std.~dev over the 32.5\,mm between taxel 1 and 6. 
        \fm{C-II}: Super-resolution factor depending on the contact force magnitude as achieved by the machine-learning solution and as predicted by the theory.
        \fm{D}: Spatially resolved position error (\fm{I}) and force error (\fm{II}). The orange lines are the TVIs for the smallest sensor value (500\,Pa in \fm{B-II}).
        \fm{E}: TVIs (\fm{I}) and spatially resolved position error (\fm{II}) for another 1D sensor layout with varied adjunct sensor distances.
    }\label{fig:Theory1D-R}
\end{figure}

\paragraph{TVIs of real sensors}
We consider strain gauges that measure the change in curvature along one direction averaged over the sensing area~\cite{SG1,SG2};
accelerometers that are able to measure the absolute inclination of the local elastomer patch using the gravity direction as a reference~\cite{Accmeter}; 
and barometers that sense the volume change caused by the material's deformation in the form of isotropic pressure~\cite{TakkTile,BaroHand}.
As detailed in~\sec{sec:sup:sensortypes}, the strain gauge has a non-monotonic behavior where one strain gauge value has several position-force possibilities.
The accelerometer does not have this problem but has a tiny ``blind spot'' directly above the sensor unit (no inclination), however super-resolution localization is well possible.
The barometer shows the convex and monotonic properties as described in our proposed theory.

\paragraph{Quantitative analysis in 1D}
Due to the convexity and symmetry of the barometer's isolines, we use it to validate our proposed theory.
We mold six barometers in the elastomer along a straight line with approximately 6.5\,mm distance to each other as shown in~\figS{fig:Theory1D-R}{A}.
The testbed carries a 4\,mm spherical indenter, and contacts the surface along the sensor placement center line.
Sensor values, force values, and indentation positions and depths are recorded at 2501 positions evenly along the line (50\,mm in total) with 40 incremental indentation depths (0.1\,mm each) at each position.
The resulting taxel responses and the TVIs are presented in~\fig{fig:Theory1D-R}{B}.
They suggest a very good super-resolution potential.

As mentioned above, the measurement noise ($\sigma_S$) in the sensor units introduces uncertainties in the position and force strength inference ($\sigma_P$ and $\sigma_F$).
To solve the inverse problem of predicting the indentation position and force magnitude from the sensor measurement, we employ a machine learning model using squared error loss, which yields a prediction with minimal variance.
In this way, we circumvent a manual computation of intersection areas, which would pose problems with real-world deviations from the idealized TVIs. 
Nevertheless, our theory can be seen as an upper bound for the performance of the machine-learning performance. 
For the architecture and training details, see~\sec{sec:sup:1D_ML}.
The results are summarized in~\figS{fig:Theory1D-R}{C \& D}.
The inference accuracy of the position and force magnitude is higher with stronger indentation force, but generally very accurate (evaluated at locations that were not included during training), as shown in~\figS{fig:Theory1D-R}{C-I}.
\FigS{fig:Theory1D-R}{C-II} shows the averaged super-resolution factors $\Omega$ (\eqn{eqn:superresfactor}) predicted by our theoretical analysis and achieved by the machine learning solution, which have a remarkable correspondence.
As expected, the theoretical predictions are higher, as they represent the best achievable result.
Higher force allows for higher super-resolution.
Averaging it over the applied force range (from 2\,mN to 1.5\,N), we obtain average super-resolution factors of 109 with the ML method compared to the prediction of 165 from the theory.
The spatial resolution of position error and force magnitude error of the prediction models are shown in~\fig{fig:Theory1D-R}{D}.
To show the impact of different TVI overlaps, we present in~\figS{fig:Theory1D-R}{E-I} a sensor with varying distance between the sensor units.
\FigS{fig:Theory1D-R}{E-II} shows the spatial resolution of position error thereof, which reveals the direct effect of the distance between adjunct sensor units.
In both cases, the overall shape resembles our theoretical prediction, \cf,~\fig{fig:Theory1D}: higher errors occur in locations where less TVIs overlap.

\subsection{Super-resolution in 2D}
\begin{figure}[tbp]
    \centering
    \includegraphics[width=\columnwidth]{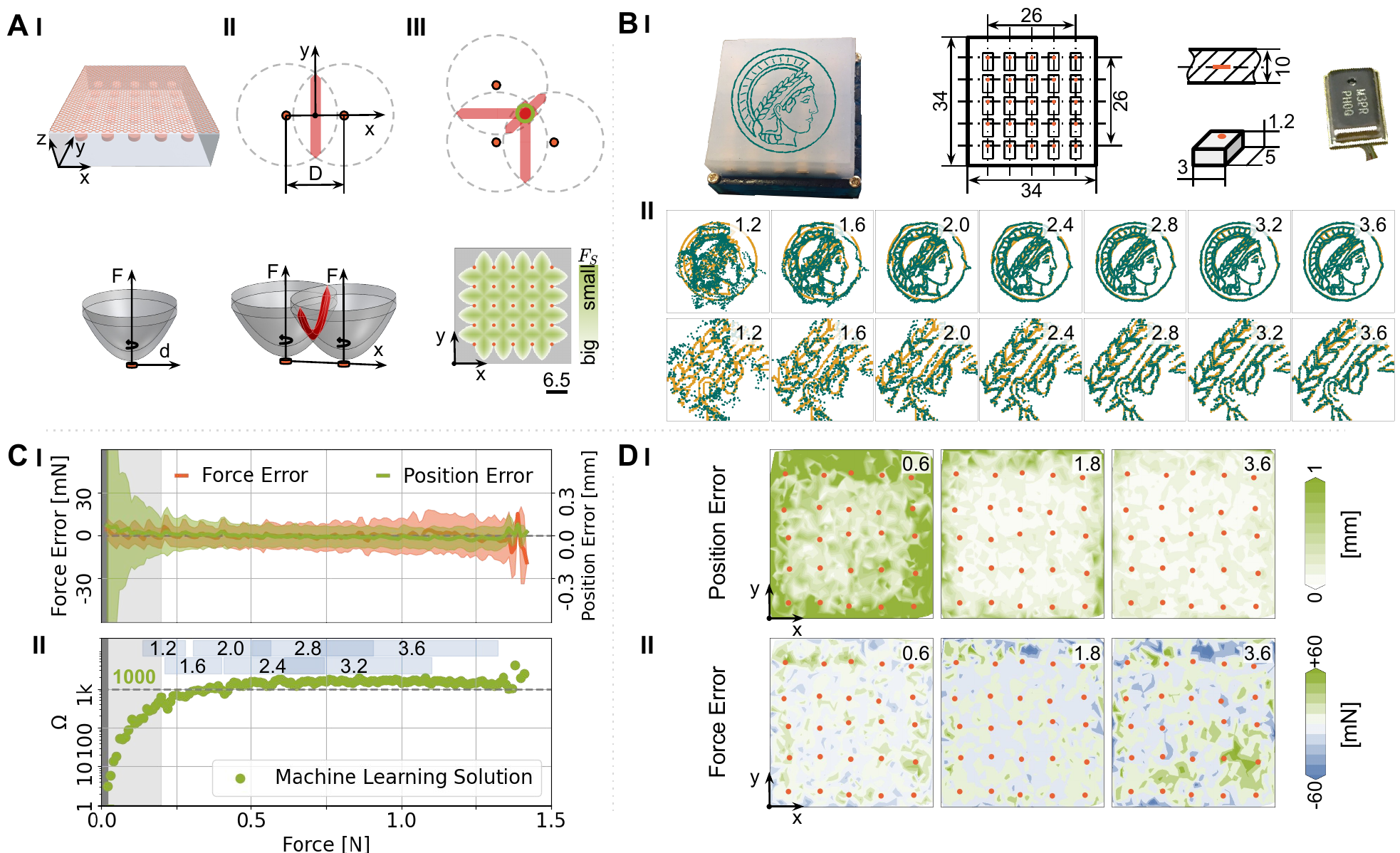}\\[-.8em]
    \caption{Super-resolution in 2D.
    \fm{A}: 2D sensor arrangement and taxel values isolines for a 2D sensing surface.
    \fm{A-I bottom}: Taxel value isoline (iso-surface) for a single taxel.
    \fm{A-II}: Intersection volume (due to measurement uncertainty) for two taxels at a distance D along the $x$ axis, see top view.
    The localization would be very uncertain along the $y$ direction.
    \fm{A-III}: Proper localization requires at least three taxels for one contact point.
    The lower part shows a grid sensor arrangement with resulting sensitivity over the surface.
    Locations between taxels are more sensitive, requiring a smaller force ($F_S$) to activate them.
    \fm{B} shows a validation of our theory on a customized 2D sensor.
    \fm{B-I} presents a picture of the sensor together with its geometric properties in a millimeter (mm) scale.
    \fm{B-II} shows a performance validation for the super-resolution functionality.
    A 4\,mm spherical indenter contacts the surface following a circular "Minerva" pattern with different indentation depths (yellow), and a trained machine learning model infers contact locations (green).
    The comparison between ground truths (yellow) and inferences (green) are visualized \wrt indentation depths (indicated at the top right corners).
    \fm{C, D}: Quantitative evaluation of the sensor device using machine learning models to infer the position and force magnitude.
    \fm{C-I}: Position and force errors depending on stimulation forces: mean and std.\ dev.\ over the sensor center region (26\,mm $\times$ 26\,mm).
    \fm{C-II}: Super-resolution factor depending on contact force magnitudes.
    \fm{D}: Spatially resolved position error (\fm{I}) and force error (\fm{II}) \wrt indentation depths.
    }\label{fig:Theory2D}
\end{figure}

The above analysis for the 1D case helps us to investigate a sensor with a flat or curved 2D sensing surface.
To simplify the analysis, we continue to assume a homogeneous transmission medium and an isotropic sensor unit, which is, for instance, approximately true for barometric sensors but violated for strain-gauge sensors.
We consider a flat 2D sensing surface with coordinates $x$ and $y$.
The concept of isolines translates into isosurfaces, as shown in~\figS{fig:Theory2D}{A-I}.
However, we still call them TVIs for consistency.
Clearly, with only two taxels, an accurate localization can only be done along one dimension ($x$) but not along two ($x$ and $y$), as shown in~\figS{fig:Theory2D}{A-II}.
To make proper super-resolution localization possible, at least three taxels need to respond to a stimulus, as shown in~\figS{fig:Theory2D}{A-III}.
As before, the accuracy increases if more taxels are involved, and the sensitivity is higher between taxels over the sensing surface (\figS{fig:Theory2D}{A-III}).
As detailed in~\sec{sec:sup:Theory2D}, to localize two simultaneous contacts, at least 6 taxels are required to respond.
Similar to the 1D case, if the contacts are too close, spurious intersections can occur. 
The spurious intersections can be ruled out because of high elongation in one direction, similar to~\figS{fig:Theory2D}{A-II}.
This is a new feature that was not observed in 1D.
Very similar considerations are also valid for curved sensing surfaces, which is illustrated in~\figS{fig:sup:2D-TheoryDemo}{C-IV}.

We mold 25 barometers in the elastomer with a 5 $\times$ 5 grid layout with approximately 6.5\,mm distance to each other as shown in~\figS{fig:Theory2D}{B-I}.
A testbed carries a 4\,mm spherical indenter and contacts the surface at given locations, similar to~\figS{fig:Theory1D-R}{A}.
Sensor values, force values, and indentation positions and depths are recorded at 69 $\times$ 69 positions (a grid with 0.5\,mm apart from each other) evenly distributed on the sensing surface (34\,mm $\times$ 34\,mm) with 20 incremental indentation depths (0.2\,mm each) at each position.
The resulting taxel responses are presented in~\fig{fig:sup:2D-BMV}.

We employ two machine learning models to predict the indentation position and force magnitude, respectively.
The architecture and training details are provided in~\sec{sec:sup:Theory2D}.
The results are summarized in~\figS{fig:Theory2D}{C \& D}.
The inference accuracy of the position and force magnitude is higher with stronger indentation force, as shown in~\figS{fig:Theory2D}{C-I}.
\FigS{fig:Theory2D}{C-II} shows the averaged super-resolution factor $\Omega$ of 1\,254 over a force range (from 2\,mN to 1.4\,N).
The averaged errors over a force range (from 0.2\,N to 1.4\,N) for locations and force magnitudes are very accurate.
Contact positions are located with an average error of 0.085\,mm (RMSE) and the contact force is estimated as correct up to 0.01\,N (RMSE).
The spatial resolution of position and force magnitude error \wrt indentation depth are shown in~\figS{fig:Theory2D}{D}.
With increased indentation depth, the position accuracy improves. 
In coherence with our theory, at the boundaries this improvement is smaller because less TVIs overlap.
In comparison, the spatial distribution of the force accuracy is relatively homogeneous.
The accuracy decreases slightly with increased indentation depth, and the boundaries tend to have an underestimation of force magnitudes.
Notice, our testbed has a position precision of 0.05\,mm, the force-torque sensor has a force precision of 0.01/0.01/0.02\,N ($F_x/F_y/F_z$).

\section{Discussion}
We present a new way to characterize, analyze, and predict force sensation at super-resolution for haptic sensors.
Our theory is based on sensor isolines that allow for a direct assessment of the uniqueness of contact position reconstruction. 
We derive quantities such as minimum force sensitivity, localization, and force sensing accuracy.
These allow us to analytically compute the super-resolution factor, namely, the number of distinguishable locations between two real sensor units.
With the help of an FEM model, we give guidelines for common design choices, such as placement of the sensor units within an elastomer as well as material properties.
We analyze three commonly used sensor types: strain gauges, accelerometers, and barometers.
Both, accelerometers and barometers, show the necessary properties for super-resolution sensation.
We conduct two cases study to evaluate our theory using a line (1D) and a grid (2D) of barometer sensor units embedded in elastomer skin.
Following our theoretically derived design choices enables us to achieve a remarkable performance of 109-fold/1\,254-fold super-resolution for the 1D and the 2D layout using machine learning methods.
Our theory predicts a super-resolution factor of up to 165 for the 1D case.

We hope that our approach can help the design of new and capable haptic sensors.
The major insights from our study are:
The sensor units in the transmission medium (elastomer) should have convex isolines.
It is beneficial to have the sensor units ``float'' in the center of the elastomer or closer to the sensing surface.
Flexible wiring helps to have the best sensitivity.
A thicker elastomer layer seems beneficial, and materials with small Young's modulus, high Poisson's ratio, and big yield strength are recommended.
The distance between sensor units should be such that for a majority of forces, the isolines of neighboring taxels intersect.
For inferring simultaneous contacts, a single sensor unit should be only activated by a single contact, and the distinguishable distance increases when the force magnitudes of multiple contacts are higher.

The actual information inference can be implemented using machine learning, as we present here.  
The decisive quantity for super-resolution perception is spatial distribution of the uncertainty of contact location and force magnitude inference. 
Exploiting the equivalence of minimizing model uncertainty by maximum likelihood estimation and least-squares error minimization for Gaussian residuals, the machine learning model directly optimizes for the quantities of interest.
Whereas the theory describes an upper limit for the super-resolution capabilities under the model-assumptions, we find that the ML methods approach the predicted accuracy in the considered real-world sensors.
In this way, our theory offers guidelines towards the system-level design of machine-learning-driven haptic sensors.
Besides the above-mentioned mechano-electrical properties, the predictions by the theory can be used to validate the suitability of the employed ML model, the data collection, and the training procedure, as we can expect the model and the predictions to be aligned. 
For instance, inaccuracies in the testbed or recording a lot of data in regions without TVI overlaps can lead to suboptimal overall performance. 

Future work will be devoted to analyze the accelerometers in more detail and investigate shear forces.
Furthermore, structured transmission media \eg with ridges~\cite{Ridges} and with multiple layers~\cite{e-Skin} are an interesting direction.

\section{Methods}
We conducted several experiments to make informed design choices and validate the proposed theory.

\paragraph{Finite element analysis}
To analyze the physical factors of the transmission medium influencing TVIs, we built a suitable finite-element model using Ansys~\cite{ANSYS}.
The model includes suitable default values of Young's modulus and Poisson's ratio for the material of EcoFlex 00-30 (70\,kPa, 0.4999).
Details of the modeling can be found in~\sec{sec:sup:fem}.

\paragraph{Sensor}
We design two sensors (1D and 2D) to validate the proposed theory.
The 1D sensor embeds six barometric units (MPL3115A2) along a line (6.5\,mm spacing) inside an elastomer (Diameter: 120\,mm, Thickness: 10\,mm), and the 2D sensor comprises 25 barometric units with a 5 $\times$ 5 grid layout (6.5\,mm spacing) inside an elastomer (Dimension: 34\,mm $\times$ 34\,mm, Thickness: 10\,mm).
We solder extra thin wires (CU-enameled wire with a diameter of 0.15\,mm, ME-Me\ss Systeme GmbH) to the units to be able to model them inside the elastomer (EcoFlex 00-30) with minimal mechanical influence of the taxel.
We design molds for the sensor spacing and the elastomer molding procedure.
The molds are 3D-printed (3D printer: Formlabs Form 3, Material: Tough).
We mold these barometric units floating in the middle of the elastomer.
The sensor values of the barometric units (MPL3115A2) are acquired through the evaluation board supplied by the Adafruit with additional 16-channeled analog multiplexers (CD74HC4067); and all of them are delivered to a laptop (ThinkPad L570) through an Arduino Mega 2560.

\paragraph{Testbed}
We create a custom testbed with three degrees of freedom (DoF).
Three DoF control the Cartesian movement of the probe ($\vec{x}, \vec{y}, \vec{z}$) using linear guide rails (Barch Motion) with a precision of $0.05\,$mm.
The 4\,mm spherical probe is fabricated from an aluminum alloy and is rigidly attached to the Cartesian gantry via an ATI Mini40 force/torque sensor with a force precision of $0.01/0.01/0.02\,$N ($F_x/F_y/F_z$).
The sensor is positioned on the base, and the indenter is used to contact it at the desired location.

\paragraph{Data}
To obtain a variety of normal forces, the indenter is moved to a specified location, touches the sensing surface, deforms it increasingly by moving normal to the surface with fixed steps of 0.2\,mm.
After a pause of 2 seconds to allow transients to dissipate, we simultaneously record the contact location, the indenter contact force from the testbed's force sensor, and the barometric units' values from the molded sensors.
All the data are collected and combined using a standard laptop.
More specifically, for the 1D sensor (\fig{fig:Theory1D-R}), the testbed makes the indenter contact 2501 positions evenly spread along the sensor center line (from -25\,mm to 25\,mm) with 40 incremental indentation depths (0.1\,mm each) at each position.
For the 2D sensor (\fig{fig:Theory2D}), 69 $\times$ 69 positions (a grid with 0.5\,mm apart from each other) evenly distributed on the sensing surface (34\,mm $\times$ 34\,mm) are probed by the indenter with 20 incremental indentation depths(0.2\,mm each).
The data of the "Minerva" pattern (Diameter: 26\,mm) is collected similarly for validating the super-resolution functionality.
The pattern has a position resolution of 0.007\,mm.
The indenter probes the sensor at each position with 10 incremental steps (0.4\,mm each with a transients pause of 2 seconds).

\paragraph{TVIs and related parameters}
We implement the following steps to compute the isolines (\fig{fig:Theory1D-R} and \fig{fig:sup:real-sensors}).
First, we linearly interpolate the sensor values and force values.
Second, we choose a position-related sensor value and find the corresponding force measurement in that position.
Third, we draw the position-force curve for that sensor value with the same color as shown in~\fig{fig:Theory1D-R}{B \& E}.
Based on the derived isolines, we use curve function ($\lambda \cdot d^\alpha+g(S)$) to fit each derived isoline.
The constant $c$ is calculated by dividing $g(S)$ over $S$.
The super-resolution factor $\Omega$ is calculated thereby using~\eqn{eqn:superresfactordetail}.

\paragraph{Machine learning}
For the 1D sensor, we use a standard MLP (multi-layer-perceptron) with six fully connected hidden layers with 100 neurons each.
The data consist of 50\,k samples that are split into datasets of training, validation, and test with a ratio of 3:1:1.
The data within the region of from -16.25\,mm to 16.25\,mm are selected as training data from the training dataset, and this region spans exactly from the center of the left barometric unit to the right one.
The models are trained with the squared error loss, the Adam optimizer (learning rate: $5\cdot 10^{-4}$, epsilon: $10^{-5}$), and a batch size of 200 samples in 1 million iterations.
The models for position and force inferences are separately trained using the same architecture and training settings.
For the 2D sensor, we use an MLP with ten fully connected hidden layers with 100 neurons each.
The data consists of 95\,k samples that are split into datasets of training, validation, and test with a ratio of 3:1:1.
The models are trained with the \emph{L2} loss, the Adam optimizer (learning rate: $2\cdot 10^{-4}$, epsilon: $10^{-5}$), and a batch size of 100 samples in 1 million iterations.
The models for position and force inferences are separately trained using the same architecture and training settings.

\section*{Acknowledgments}
\textbf{Funding:}
The authors thank the China Scholarship Council (CSC) and the International Max Planck Research School for Intelligent Systems (IMPRS-IS) for supporting H.S.
G.M.\ is a member of the Machine Learning Cluster of Excellence, funded by the Deutsche Forschungsgemeinschaft (DFG, German Research Foundation) under Germany’s Excellence Strategy –
EXC number 2064/1 – Project number 390727645.
We acknowledge the support from the German Federal Ministry of Education and Research (BMBF) through the Tübingen AI Center (FKZ: 01IS18039B).
\textbf{Author contributions:}
H.S. and G.M. conceived the method and the experiments, drafted the manuscript, and revised it.
H.S. designed and constructed the hardware, developed fabrication methods, designed and conducted experiments, collected and analyzed the data.
G.M. supervised the data analysis.
We thank Anna Levina for discussions on the analytic solution for $\sigma_P$.

\bibliographystyle{IEEE}

\clearpage
\title{\centerline{\Huge{\textbf{Supplementary Materials for}}} \vspace{2em}
 \centerline{\Large{Theory and Design of Super-resolution Haptic Skins}}}

\renewcommand{\thetable}{S\arabic{table}}
\renewcommand{\thefigure}{S\arabic{figure}}
\renewcommand{\theequation}{S\arabic{equation}}
\setcounter{table}{0}
\setcounter{figure}{0}
\setcounter{equation}{0}
\setcounter{page}{1}
\renewcommand{\thepage}{\roman{page}}

\appendix

\paragraph{The PDF file includes:}
\newcommand{\frefitem}[1]{\item[Page \pageref{#1}:] \fig{#1}}
\newcommand{\trefitem}[1]{\item[Page \pageref{#1}:] \tab{#1}}
\newcommand{\prefitem}[1]{\item[Page \pageref{#1}:]}

\begin{figure}[tb]
    \centering
    \includegraphics[width=\columnwidth]{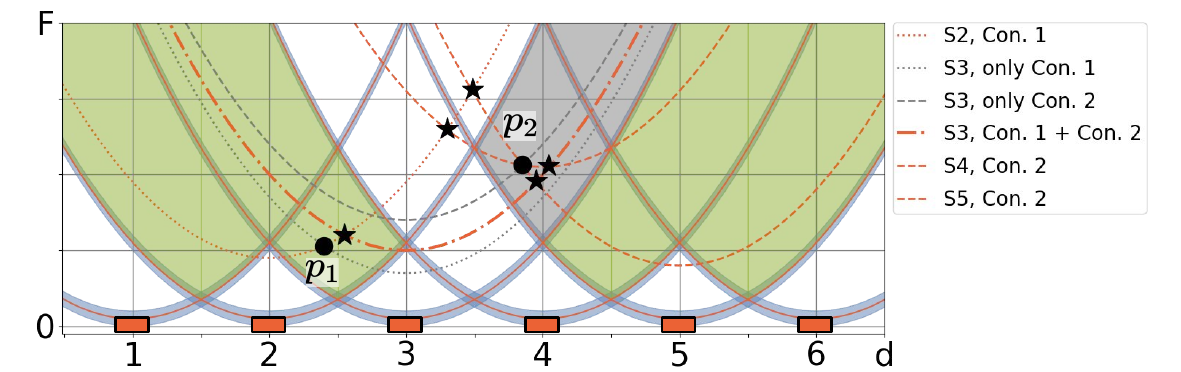}\\[-.8em]
    \caption{Multiple contacts discrimination.
    There are six sensing units aligning in one line forming a 1D layout to discriminate double contacts.
    When one contact locates in the left green-shaded area, all contact situations of the second contact can be successfully discriminated when they are in the right green-shaded area. 
    In contrast, when the second point is too close, say in the gray-shaded area, the following problem occurs. 
    The first contact at $p_1$ alone would activate two sensing units (S-2: orange-dotted line, S-3: gray-dotted line).
    Another contact in the gray-shaded area at $p_2$ would activate three sensing units (S-3: gray-dashed line, S-4: orange-dashed line, S-5: orange-dashed line).
    When these two contacts happening simultaneously, the value of the sensing unit S-3 turns to be an orange-dash-dotted line (combined TVI).
    This will result in an unreliable inference.
    Multiple spurious contact points occur ("$\bigstar$") rather than two actual contact points ("$\bullet$") }
    \label{fig:sup:Multicontact}
\end{figure}

\begin{description}[itemsep=0em,labelwidth=5em]
    \prefitem{sec:sup:computation of variance} \sec{sec:sup:computation of variance}: Computation of $\sigma_P$ and $\sigma_F$.
    \frefitem{fig:sup:sigmas} Computation of $\sigma_P$ and $\sigma_F$.
    \prefitem{sec:sup:Multicontact} \sec{sec:sup:Multicontact}: Multiple simultaneous contacts discrimination.
    \frefitem{fig:sup:Multicontact} Multiple contacts discrimination.
    \prefitem{sec:sup:fem} \sec{sec:sup:fem}: Physical factors influencing TVIs.
    \frefitem{fig:sup:fem} Physical factors influencing TVIs using FEM.
    \prefitem{sec:sup:sensortypes} \sec{sec:sup:sensortypes}: Taxel value isolines of real sensors.
    \frefitem{fig:sup:real-sensors} Response and TVIs of real sensor units.
    \prefitem{sec:sup:1D_ML} \sec{sec:sup:1D_ML}: 1D sensor.
    \frefitem{fig:sup:real-sensor-calculation} Quantitative parameters of the 1D sensor.
    \prefitem{sec:sup:Theory2D} \sec{sec:sup:Theory2D}: 2D sensor.
    \frefitem{fig:sup:2D-TheoryDemo} Taxel values isolines for a 2D sensing surface.
    \frefitem{fig:sup:2D-BMV} Response and TVIs for real sensor units on a 2D sensing surface.
\end{description}


\newpage
\section{Computation of $\sigma_P$ and $\sigma_F$}~\label{sec:sup:computation of variance}
\begin{figure}[H]
    \centering
    \includegraphics{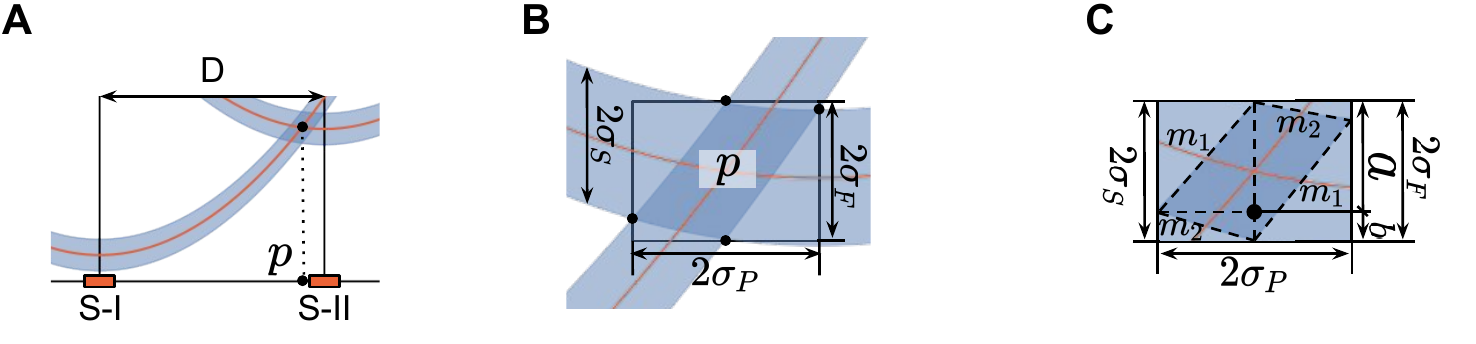}
    \caption{Computation of $\sigma_P$ and $\sigma_F$.
    \fm{A} shows the intersection of two TVIs caused by a single contact at position $p$.
    \fm{B}: Precise computation of $\sigma_P$ and $\sigma_F$ based on the intersection type. See also \FigS{fig:Theory1D}{A} for different intersection types.
    \fm{C}: Approximated computation of $\sigma_P$.
    The intersection area of two TVIs between the taxels can be well approximated by a parallelogram. 
    The slopes of the edges ($m_1$, $m_2$) are simply given by the inclination of the TVIs at the intersection point (first order approximation).}
    \label{fig:sup:sigmas}
\end{figure}

We first go through the accurate computation of the inference uncertainties for the idealized TVIs that is used in the numeric results.  
Then we present an approximation that we use to compute the super-resolution factor analytically.
\paragraph{Precise computation of $\sigma_P$ and $\sigma_F$}\label{sec:sup:sigmas}
For two taxels at distance $D$, as shown in \figS{fig:sup:sigmas}{A}, and the TVIs described by the following equations:
\begin{align}
    I_1^{S_1}(d) &= g(S_1) +  \lambda |d|^{\alpha}\label{eqn:sup:TVI1},\\
    I_2^{S_2}(D-d) &= g(S_2) +  \lambda |D-d|^{\alpha}\label{eqn:sup:TVI2},
\end{align}
we use the exact corners of the intersection area of the two TVIs with their measurement uncertainty hose.
These intersection points are computed numerically, and we use this method for all plots in the main paper.

In more detail, for each TVI we have an upper and lower bound for the uncertainty (one standard deviation) denoted by $h^\pm$:
\begin{align}
    h^+_1 &= I_1^{S_1}(d) + \sigma_S\\
    h^-_1 &= I_1^{S_1}(d) - \sigma_S\\
    h^+_2 &= I_2^{S_2}(D-d) + \sigma_S\\
    h^-_2 &= I_2^{S_2}(D-d) - \sigma_S.
\end{align}
There are 4 intersection points: $h^+_1 \cap\, h^+_2$, $h^+_1 \cap\, h^-_2$, $h^-_1 \cap\, h^+_2$, and $h^-_1 \cap\,  h^-_2$ which we illustrate in \figS{fig:sup:sigmas}{B}.
The size of the bounding box of these four points defines $\sigma_P$ and $\sigma_S$.  

\paragraph{Approximate analytical computation of $\sigma_P$}\label{sec:sup:approxsigmaP}
Here, we give a closed-form expression relating the force measurement noise to the position accuracy between two taxels.
We make a first-order approximation of the intersection area as a parallelogram, as shown in \figS{fig:sup:sigmas}{C}.
For two taxels at distance $D$ and the TVIs described by \eqn{eqn:sup:TVI1} and \eqn{eqn:sup:TVI2}, we can compute the (absolute) derivatives of the TVIs at the intersection point $d$:
\begin{align}
  m_1 &:= \left|\frac{\mathrm{d} I_1^S(d)}{\mathrm{d}d}\right| = \lambda \alpha |d|^{\alpha -1 }\\
  m_2 &:= \left|\frac{\mathrm{d} I_2^S(D-d)}{\mathrm{d}d}\right| = \lambda \alpha |D-d|^{\alpha -1}.
\end{align}
The trigonometric construction is shown in
\figS{fig:sup:sigmas}{C} and yields the following result:
\begin{align}
2\sigma_S &= a+b  = m_1 \sigma_P + m_2 \sigma_P\\
\sigma_P &= \frac{2\sigma_S}{m_1 + m_2}\label{eqn:sup:sigmaP-sigmaS}
\end{align}

\section{Multiple Simultaneous Contacts Discrimination}\label{sec:sup:Multicontact}
In many applications, we are interested in detecting multiple simultaneous contact points.
In this section, we investigate the criterion to discriminate double contacts.
At first glance, we need 2 pairs of taxels.
However, when the contact points are too close, spurious contact points would be detected due to additional intersections of TVIs.
The basic condition for successfully distinguishing them is shown in \fig{fig:sup:Multicontact}.
We consider two contact points with individual force strength are located in a 1D sensor layout with six sensing units.
When one contact is located in the left green-shaded area, then a second contact can be clearly discriminated when it is within the right green-shaded area.
In contrast, when the second contact is located in the gray-shaded area, it activates at least three sensing units (S-3,S-4,S-5).
Notice, the sensing unit S-3 is also activated by the first contact.
Thus, the value of S-3 is then not reliable for triangulation inference and will create spurious inference points and modify the position of the otherwise correctly inferred locations.
The most important take-home message is to have at least two taxels between these two contacts
and at most one taxel of them that is evoked by one contact not the other.
With higher forces applied by these two contacts, the distinguishable distance for double simultaneous contact is larger.

\section{Physical Factors Influencing TVIs}\label{sec:sup:fem}
We investigate the detailed influences of the properties of the transition medium (\eg elastomer) and the taxel placement on the TVIs and on the sensitivity.
External contact at the sensing surface causes deformation of the transmission medium that can be measured by physical sensor units (taxels). 
In this section, we assume the taxels are able to measure deformation/strain within the medium, either isotropically or directionally.

In principle, the deformation can be described by the absolute movement of elements, called displacement, and by the relative movement of elements, denoted as strain.
To study the isolines we consider the displacement as a measure of deformation. 
The displacement is computed for a static mechanical model simulated with the finite element method (FEM) using Ansys~\cite{ANSYS}.
The questions we want to answer in this section are: 
Where to place the sensor units within the transition medium?
How thick should be the transition medium?
What is the effect of material properties such as the Poisson's ratio?

\begin{figure*}[htp]
    \centering
    \includegraphics[width=\columnwidth]{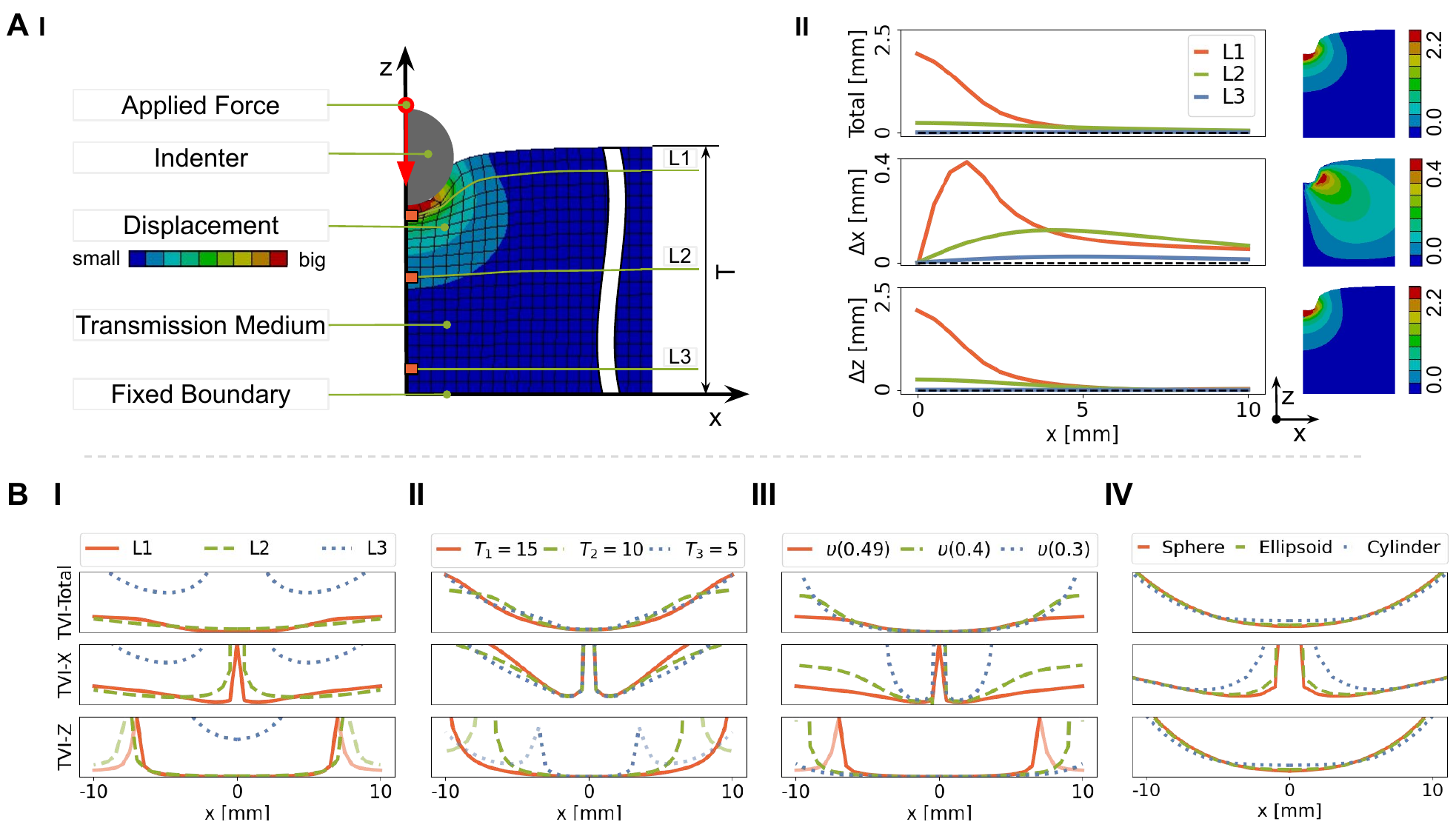}\\[1em]
    \caption{Physical Factors Influencing TVIs using FEM.
        \fm{A-I}: The model: because of symmetry, only one half is shown. 
        The thickness is $T=10$\,mm, and we consider taxels at level $L1=T-1\,$mm, $L2=T-5\,$mm and $L3=T-9\,$mm.
        \fm{A-II}: Total displacement maps as well as its $x$ and $z$ components. 
        \fm{B}: TVIs for hypothetical sensors measuring total displacement or component-wise displacement, for different taxel depth (\fm{I}), material thickness (\fm{II}), Poisson's ratios (\fm{III}), and indenter shapes (\fm{IV}).
        TVIs for negative sensor values are in a light shade.
        Default configurations are: Poisson's ratio $\nu=0.49$, Young's modulus $E=0.07\,$MPa, and density $\rho=1.07\,$g/cm$^3$ for the transmission medium;
        Poisson's ratio $\nu=0.33$, Young's modulus $E=71\,$GPa, and density $\rho=2.77\,$g/cm$^3$ for the indenter; bonded contact type without friction consideration.
        }
    \label{fig:sup:fem}
\end{figure*}

\paragraph{Simulation model}\label{sec:sup:fem-model}
The full model is a cylinder-shaped sensor transmission medium with a diameter of 200\,mm and a thickness of 5\,mm, 10\,mm,  or 15\,mm,
but we simplify the model in the FEM simulation into a 2D plane due to the axis-symmetric property of the cylinder and the spherical indenter (10\,mm diameter) as shown in~\figS{fig:sup:fem}{A-I}.
The transmission medium is constrained at the bottom by a fixed boundary. 
We analyze the displacement when a normal force is applied to the contact surface via the indenter.
In \figS{fig:sup:fem}{A-II} the displacement is evaluated in $x$- and $z$-direction, as well as in terms of total displacement.
Considering the displacement at three different depths, close to the sensing surface (L1), in the middle (L2), and close to the fixed boundary (L3), shows significant differences.
The displacement in depth direction ($\Delta z$) decreases monotonously with distance, but is much stronger at L1 than at L3.
The radial displacement ($\Delta x$) first increases and then attenuates, which will give rise to non-monotonous TVIs.

\paragraph{Taxel depth}
The dependency of the deformation on the depth has an impact on the TVIs.
\FigS{fig:sup:fem}{B-I} shows the TVIs for ideal displacement sensors at three different depths (L1-L3) described above. 
We plot the TVI for the smallest non-zero taxel value. 
However, because there can also be negative displacements, we add a light-shaded TVI for small negative values if occurring. 
When measuring depth deformation, the TVIs have a convex shape for positive sensor values and a second parabola at either side for the negative sensor values. 
This holds for all depths, although only visible for L1 due to a limited plot range.
The radial direction is more surprising, as the center is not moving radially, the TVIs have a pole in the center.
The total deformation is monotonous for low to middle taxel depths (L1 and L2).
Depending on the physical quantity measured by the real sensor unit, this leads to different conclusions:
For strain gauges and accelerometers, for instance, the global displacement can be considered as a good proxy, as they measure the curvature and inclination of the material. 
Thus, these sensors should be more close to the sensing surface.
Barometric sensors, for instance, measure local displacement, for which we cannot give a recommendation at this point.

\paragraph{Sensing surface thickness}
Guided by the previous result, we consider a taxel at depth L1 (1\,mm below the sensing surface) for different total thicknesses of the elastomer.
\FigS{fig:sup:fem}{B-II} presents the TVIs. 
The radial deformation is, as expected, only a little effected by the amount of material underneath the taxel.
The depth deformation, on the other hand, shows a much larger effect of thickness, because for smaller thickness we get less displacement. 
In summary, the thicker material causes more displacement and in turn increases the sensitivity, because the fixed boundary becomes less dominant.
Notice, for thicker material, the shear/x-directional displacement has stronger attenuation property and has less shear sensitivity.
Nevertheless, there are several ways to increase the sensitivity in the shear direction, as proposed in earlier work~\cite{Ridges,e-Skin}.
One is to add ridges on the contact surface~\cite{Ridges}.
Another is to use multiple layers of materials to cause mechanical contrast and affects the force distribution~\cite{e-Skin}.

\paragraph{Material properties}
For a robust sensor design, we need to make sure that the contact-caused stress is smaller than the yield strength of the transmission medium, and the material deformation is an elastic behavior.
The main two properties of the transmission material we consider are the Young's modulus and the Poisson's ratio.
The Young's modulus describes how easy the material deforms and has a proportional impact on the deformation.
A soft material (small Young's modulus) improves sensitivity, but is also increasingly deformed by inertial effects.
Poisson's ratio measures the relative transversal/radial expansion when the material is axially compressed.
Most elastomers have Poisson's ratio around 0.5 and metals have around 0.3~\cite{PoissonRatio}.
\FigS{fig:sup:fem}{B-III} shows the TVIs for different Poisson's ratios.
Decreasing the ratio, the radial displacement becomes much less sensitive, whereas the depth displacement has higher sensitivity (lower TVIs).
Thus, depending on the measurement direction of the real sensor unit, different Poisson's ratios are preferred. 

\paragraph{Indenter shape}
What is the impact of the indenter shape (or the object getting in touch with the sensing device later) on the TVIs?
For illustration, we simulate the displacement caused by three different indenter shapes: a sphere, an ellipsoid, and a cylinder.
The displacement decreases and becomes flatter for the cylinder near the contact center, as shown in \figS{fig:sup:fem}{B-IV}.
The cylinder with flatter contact surface causes less sensitivity for shear displacement.
Increasing the indenter size, the displacement becomes smaller, which results in less sensitivity too.
As a remark, for too small indenters, there is a risk to exceed the material yield strength and break the contact surface material.

\begin{figure*}
    \centering
    \includegraphics[width=\columnwidth]{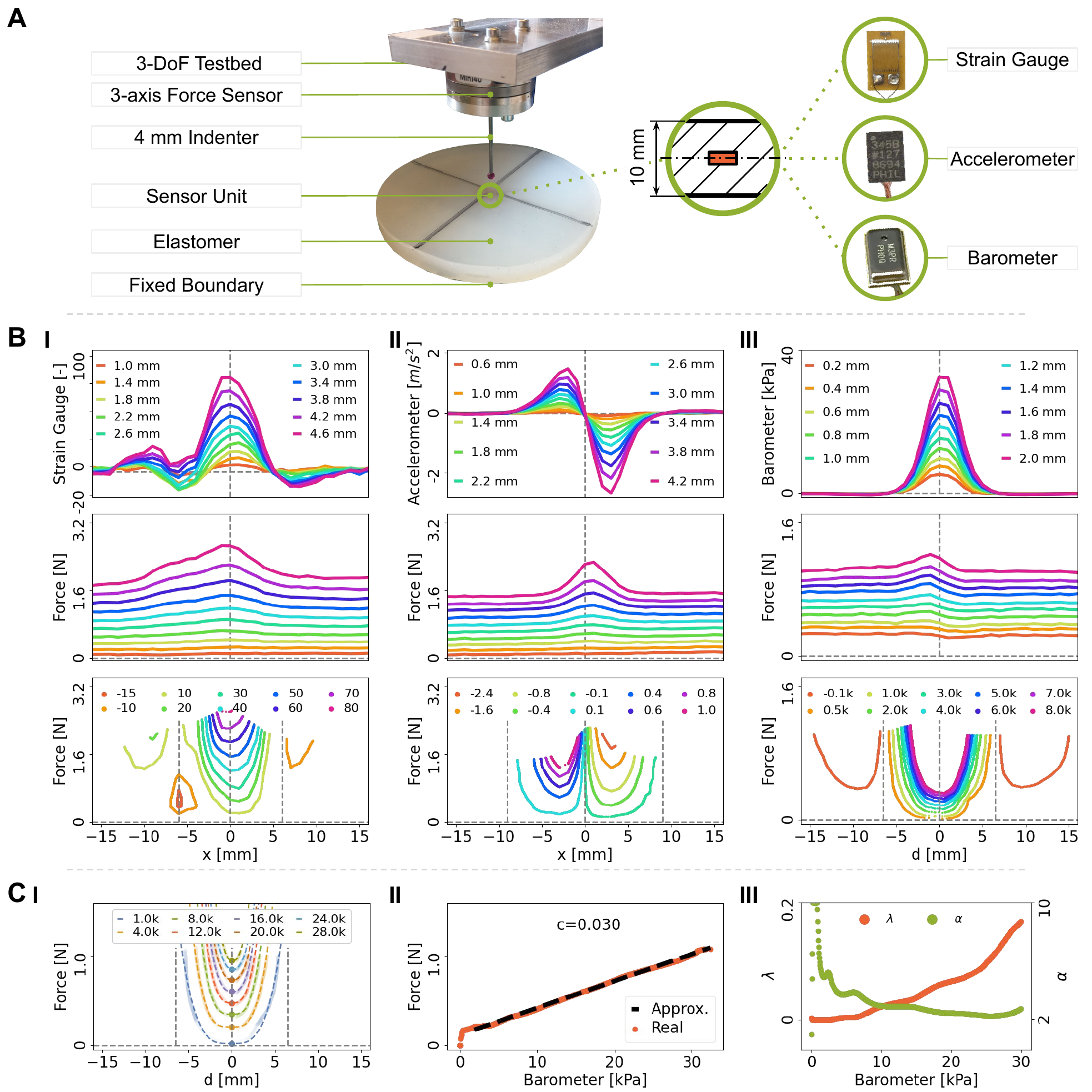}\\[-.5em]
    \caption{Response and TVIs of real sensor units.
    \fm{A}:~The experimental setup.
    The black lines on the disc mark the stimulation points.
    \fm{B}:~Sensor response (top), true indentation force (middle), and TVIs (bottom) for strain gauge (left), accelerometer (center), and barometer (right).
    Lines are presented for different penetration depths of the indenter and different sensor values in the bottom row.
    Barometers have TVIs that are most similar to our theoretical model.
    \fm{C}:~Analytical approximation of barometric TVIs \eqnp{eqn:sup:curve-fitting}: isoline shape, constant $c$ at $f(F,0)$, coefficient $\lambda$ and attenuation power $\alpha$.
    }
    \label{fig:sup:real-sensors}
\end{figure*}

\section{Taxel Value Isolines of Real Sensors}\label{sec:sup:sensortypes}
We complement our theoretical analysis by measuring the response curves of three suitable sensors.
We consider strain gauges that measure the change in curvature along one direction averaged over the sensing area~\cite{SG1,SG2};
accelerometers that are able to measure the absolute inclination of the local elastomer patch using the gravity direction as a reference~\cite{Accmeter};
and barometers that sense the volume change caused by the material's deformation in the form of isotropic pressure~\cite{TakkTile,BaroHand}.

\paragraph{Experimental setup}
As shown in \figS{fig:sup:real-sensors}{A}, we mold a single sensor unit at the center of an elastomer disc (120\,mm diameter and 10\,mm thickness) and measure the sensor unit's response when indenting it with a controlled depth along two perpendicular axes using an automated 3-DoF testbed with a force sensor.
The strain gauge sensor is EP08-250AE-350, the accelerometer is ADXL345, and the barometer is MPL3115A2.
We solder extra thin wires (CU-enameled wire with a diameter of 0.15\,mm, ME-Me\ss Systeme GmbH) to the chips to be able to mold them inside the elastomer with minimal mechanical influence of the taxel.
We mold these three sensors floating in the middle and center of three elastomer pieces (EcoFlex 00-30) individually.
Additionally, we build a 3-DoF test bed (Barch Motion, precision: 0.05\,mm) equipped with a 3-axis force measurement device (ATI Mini40, precision: $0.01/0.01/0.02\,$N ($F_x/F_y/F_z$)) to indent the elastomer surface in a precise and automated way.

\paragraph{Data collection}
As shown in~\figS{fig:sup:real-sensors}{A}, a 4\,mm spherical indenter goes along two black-colored perpendicular trajectories where each sensor unit is located under the cross point.
The testbed is used to make the indenter contact 101 positions evenly spread along each trajectory (from -50\,mm to 50\,mm) with 25 incremental indentation depths (0.2\,mm each) at each position.
The sensor value of the strain gauge (EP08-250AE-350) is acquired through a quarter Wheatstone bridge and amplified by an MCP609;
the sensor values of the accelerometer (ADXL345) and the barometer (MPL3115A2) are acquired through the evaluation boards supplied by Adafruit;
and all of them are delivered to a personal computer through an Arduino Mega 2560.
The recorded data are sensor values, force values, and indentation positions and depths.
For different indentation depths, the sensor values vary along the indentation positions, and deeper indentations need higher forces and result in higher sensor values.
When the indention positions are near the placed sensor units, the applied forces are higher because the physical sensor unit is stiffer than the elastomer.

\paragraph{Data processing}
Based on the recorded data, we present in \figS{fig:sup:real-sensors}{B} the sensor values, the applied forces, and the TVIs for each sensor type as a spatially changing quantity. 
We only show the data for one of the two directions.
For the strain gauge unit, this is along its measurement direction.
Accelerometers are interesting, because they can distinguish between both directions.
The barometer is fairly isotropic up to small deviations because of the rectangular sensor housing.
We implement the following steps to compute the isolines:
first, we linearly interpolate the sensor values and force values;
second, we choose a position-related sensor value and find the corresponding force measurement in that position;
third, we draw the position-force curve for that sensor value with the same color.
The strain gauge has a non-monotonic behavior, where one strain gauge value has several position--force possibilities.
The accelerometer does not have this problem, however, it has a ``blind spot'' directly above the sensor unit (no inclination), however, this is a tiny area.
Note that the TVIs on both sides are for different sensor values, so super-resolution localization is well possible.
We leave a theoretical analysis of these TVI shapes for future work.
The barometer shows the convex and monotonic properties as described in our proposed theory.
Notice the two orange ``wings'' which are the TVI for negative values caused by the lateral pulling force from distant contacts.

\paragraph{TVI fitting and sensor-force relationship (constant $c$)}\label{sec:sup:sensorforcerelationship}
To support the assumption for our theory introduced in Section ``The Model'', we herein introduce the analytical steps for approximating taxel-value-isolines (TVIs) from the sensor and force measurements.
We first empirically acquire the sensor isolines, and then fit a function 
\begin{equation}
    I^S(d) = g(S) + \lambda |d|^\alpha \label{eqn:sup:curve-fitting}
\end{equation}
to each individual isoline, as shown in~\figS{fig:sup:real-sensors}{C-I}.
Taking the $g(S)$ of each barometric values at $d=0$ as the force strength, we can acquire the relationship between sensor value and force strength captured by the constant $c$ (see text below \eqn{eqn:taxelvalue-noise}).
The value of $c$ is simply the ratio between the applied force $g(S)$ and the barometer value.
As shown in~\fig{fig:sup:real-sensors}{C-II}, the linearity assumption holds.
Moreover, the attenuation coefficient $\lambda$ and power $\alpha$ are also approximated.
As shown in~\fig{fig:sup:real-sensors}{C-III}, $\lambda$ becomes larger with increased barometer value, while $\alpha$ attenuates approaching $\alpha=2$.

\section{1D Sensor}\label{sec:sup:1D_ML}
\begin{figure*}[tbp]
    \centering
    \includegraphics[width=\columnwidth]{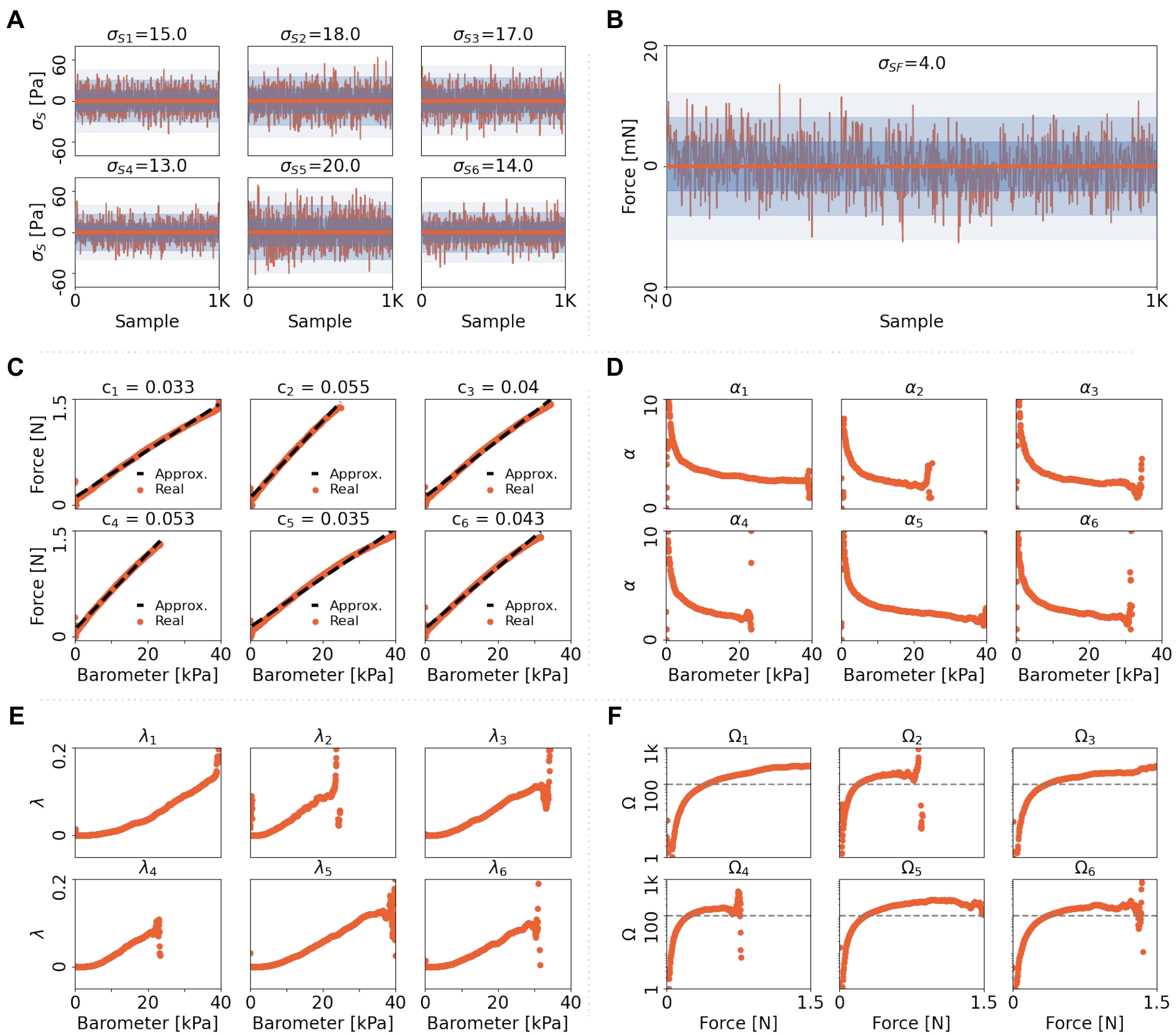}\\[-.5em]
    \caption{Quantitative parameters of the 1D sensor.
    \fm{A \& B} show the measurement noises of six barometers and the force-torque sensor.
    \fm{C}: Constant $c$ indicates the relationship between the force values and barometer values.
    Red is the real relationship, dashed blue is the linearly approximated relationship.
    \fm{D}: Attenuation power of the TVI's shape.
    \fm{E}: Attenuation coefficient of the TVI's shape.
    \fm{F}: Analytically derived super-resolution factor \wrt applied force strength.
    The dashed gray lines are with a super-resolution factor of 100. 
    }
    \label{fig:sup:real-sensor-calculation}
\end{figure*}
\paragraph{Theory}
Due to the convexity and symmetry of the barometer's isolines, we use it to validate our proposed theory.
We mold six barometers in the elastomer along a straight line with approximately 6.5\,mm distance to each other as shown in~\figS{fig:Theory1D-R}{A}.
As mentioned above, the measurement noise ($\sigma_S$) in the sensor units introduces uncertainties in the position and force strength inference ($\sigma_P$ and $\sigma_F$).
There are two major noise sources in our setting, one is the noise measurement in the six barometers $\sigma_{Si}$, another is the force-torque sensor $\sigma_{SF}$.
We collect 1000 samples of the force-torque sensor and the barometer values in the unloaded case to evaluate the measurement noise levels, as shown in~\fig{fig:sup:real-sensor-calculation}{A \& B}.
Then a testbed carries a 4\,mm spherical indenter, and contacts the surface along the sensor placement center line.
Sensor values, force values, and indentation positions and depths are recorded at 2501 positions evenly along the line (50\,mm in total) with 40 incremental indentation depths (0.1\,mm each) at each position.
We follow the steps introduced in~\sec{sec:sup:sensorforcerelationship} to acquire the constant $c_i$, attenuation power $\alpha_i$, attenuation coefficient $\lambda_i$, and use the following equation (\cf \eqn{eqn:superresfactordetail} with $D=6.5$):
\begin{equation}
    \Omega_i = \frac{{6.5\lambda_i \alpha_i} (\nicefrac{6.5}{2})^{(\alpha_i-1)}}{2 \cdot 2 (\sigma_{SF} + c_i \cdot \sigma_{Si})}
\end{equation}
to calculate the super-resolution factor $\Omega_i$ for each barometric sensor unit, assuming two identical units are used.
The overall $\Omega$ of the whole 1D sensor is 165 that is averaged over these six $\Omega_i$ for each discrete threshold of 0.02\,N over a range of (0.02\,N-1.5\,N).

\paragraph{Machine Learning}
To solve the inverse problem of predicting the indentation position and force magnitude from the sensor measurement, we employ a machine learning model using squared error loss, which yields a prediction with minimal variance.
In this way, we circumvent a manual computation of intersection areas, which would pose problems with real-world deviations from the idealized TVIs. 
Nevertheless, our theory can be seen as an upper bound for the performance of the machine-learning performance. 
We use a standard MLP (multi-layer-perceptron) with six fully connected hidden layers with 100 neurons each.
The data consist of 50\,k samples that are split into datasets of training, validation, and test with a ratio of 3:1:1.
The data within the region of from -16.25\,mm to 16.25\,mm are selected as training data from the training dataset, and this region spans exactly from the center of the left barometric unit to the right one.

\section{2D Sensor}\label{sec:sup:Theory2D}
\begin{figure*}[tbp]
    \centering
    \includegraphics[width=\columnwidth]{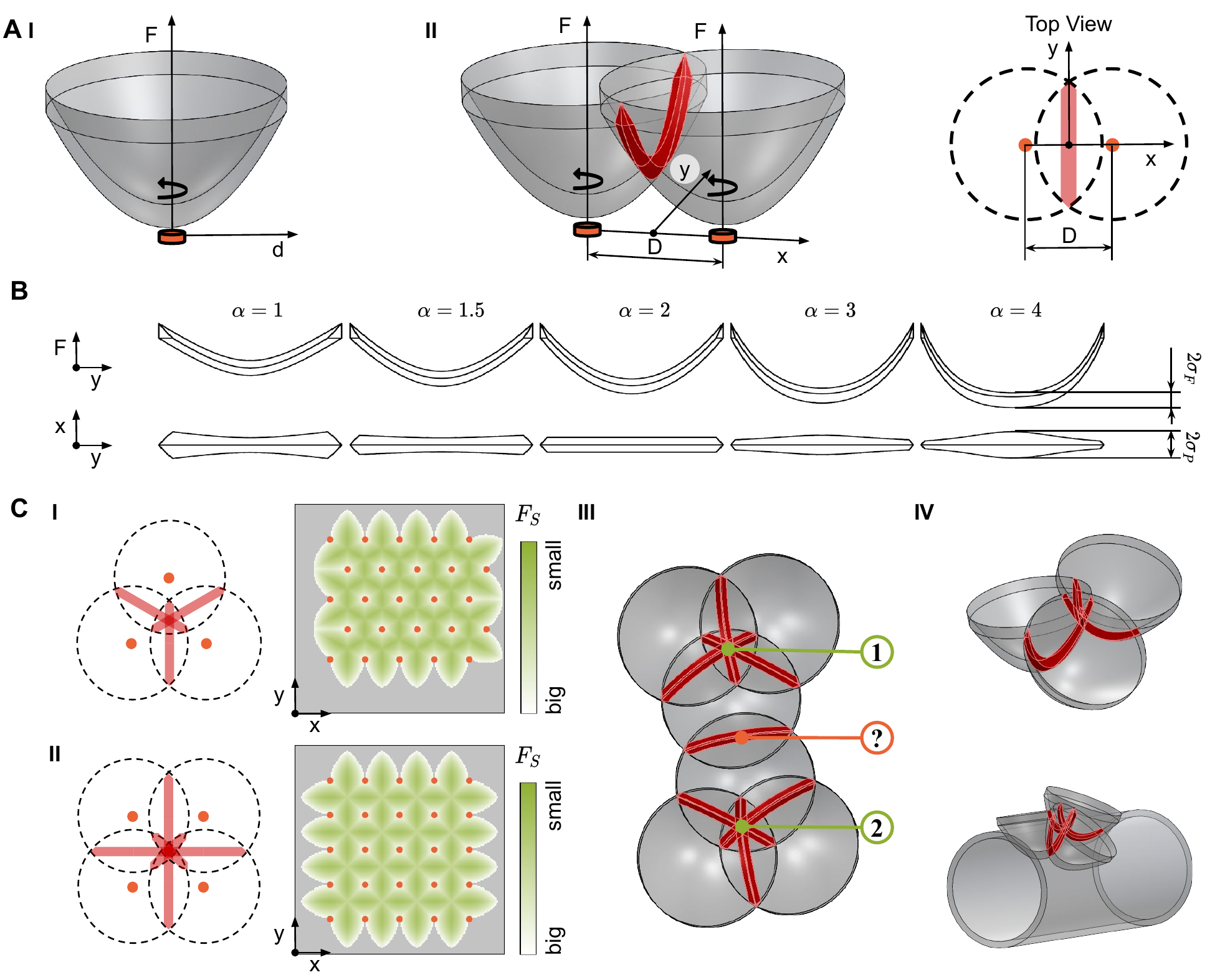}\\[-.5em]
    \caption{Taxel values isolines for a 2D sensing surface. 
    \fm{A-I}:~Taxel value isoline (iso-surface) for a single taxel.
    \fm{A-II \& III}:~Intersection volume (due to measurement uncertainty) for two taxels at a distance $D$ along the $x$ axis, see top view.
    The localization would be very uncertain along the $y$ direction.
    \fm{B}:~Intersection volume for different attenuation exponents $\alpha$.
    \fm{C}:~Proper localization requires at least 3 taxels for 1 contact point.
    \fm{I}~shows a hexagonal sensor placement with resulting sensitivity over the surface and an illustration of the intersection volume of the TVIs from 3 taxels.  
    \fm{II}~is the same for a grid and 4 taxels.
    \fm{III}~illustrates a spurious contact localization for 2 contact points and 6 taxels (marked with ``?''), however its uncertainty would be very large.
    \fm{IV}~illustrates the intersection of TVIs for the case of a curved sensing surface. 
    The same applies as in the case of planar 2D, except that distances need to measured as geodesics on the curved surface.
    }
    \label{fig:sup:2D-TheoryDemo}
\end{figure*}

\paragraph{Theory}
The above analysis for the 1-dimensional case helps us to investigate a sensor with a flat or curved 2-dimensional sensing surface.
To simplify the analysis, we continue to assume a homogeneous transmission medium and an isotropic sensor unit, which is, for instance, approximately true for barometric sensors but violated for strain-gauge sensors.
We first consider a flat 2D sensing surface with coordinates $x$ and $y$. 
The concept of isolines translates into isosurfaces, as shown in \figS{fig:sup:2D-TheoryDemo}{A-I}.
However, we still call them TVIs for consistency.
Clearly, with only two taxels, an accurate localization can only be done along one dimension, but not along two.
The intersection volume of the TVIs is presented in \figS{fig:sup:2D-TheoryDemo}{A-II \& III}. 
For attenuation exponents, $\alpha\not=2$ the intersection volume changes depending on the transversal position $y$
as shown in \figS{fig:sup:2D-TheoryDemo}{B}.
To make proper super-resolution localization possible, at least 3 taxels need to respond to a stimulus, as shown in \figS{fig:sup:2D-TheoryDemo}{C-I}. 
As before, the accuracy increases if more taxels are involved, as shown in \figS{fig:sup:2D-TheoryDemo}{C-II} for 4 taxels arranged in a regular grid.
We also compute the sensitivity distribution over the sensing surface for both sensor arrangements. 
The honeycomb pattern allows for a more homogeneous sensitivity.
To detect two simultaneous contacts, at least 6 taxels are required to respond.
Similar to the 1D case, if the contact points are too close, spurious intersections can occur.
In some cases, as shown in \figS{fig:sup:2D-TheoryDemo}{C-III}, spurious intersections can be ruled out because of high elongation in one direction. This is a new feature that was not observed in 1D. 
Very similar considerations are also valid for curved sensing surfaces, which is illustrated in \fig{fig:sup:2D-TheoryDemo}{C-IV}.

\begin{figure*}
    \centering
    \includegraphics[width=\columnwidth]{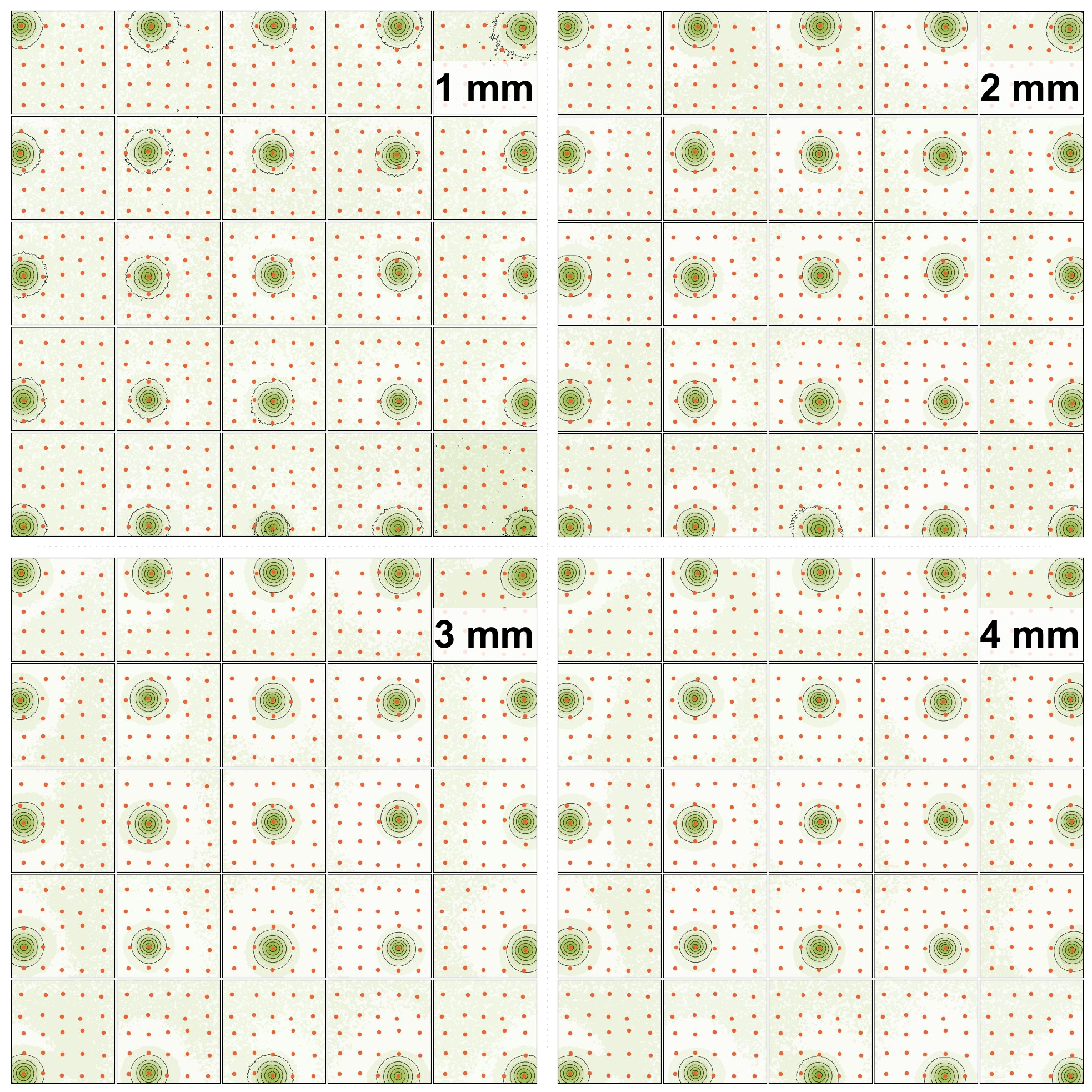}\\[-.5em]
    \caption{Response and TVIs for real sensor units on a 2D sensing surface.
    25 Barometer values for 2D sensor with different indentation depths (1\,mm, 2\,mm, 3\,mm, 4\,mm).
    Each sub-figure with the specific indention depth shows the barometer value contour maps for each of the 25 sensors marked by a red dots (center position). 
    }
    \label{fig:sup:2D-BMV}
\end{figure*}

\paragraph{Hardware and Machine Learning}
We mold 25 barometers in the elastomer with a 5 $\times$ 5 grid layout with approximately 6.5\,mm distance to each other as shown in~\figS{fig:Theory2D}{B-I}.
A testbed carries a 4\,mm spherical indenter and contacts the surface at given locations, similar to~\figS{fig:Theory1D-R}{A}.
Sensor values, force values, and indentation positions and depths are recorded at 69 $\times$ 69 positions (a grid with 0.5\,mm apart from each other) evenly distributed on the sensing surface (34\,mm $\times$ 34\,mm) with 20 incremental indentation depths (0.2\,mm each) at each position.
The resulting taxel responses are presented in~\fig{fig:sup:2D-BMV}, which show approximately isotropic properties along all directions.
We employ two machine learning models to predict the indentation position and force magnitude, respectively.
We use an MLP with ten fully connected hidden layers with 100 neurons each.
The data consists of 95\,k samples that are split into datasets of training, validation, and test with a ratio of 3:1:1.
The models for position and force inferences are separately trained using the same architecture and training settings.

\paragraph{Super-resolution Factor $\Omega$}
Using the trained machine-learning models, we evaluate the super-resolution factor of the 2D sensor as the fraction between the number of virtual taxels $n_v$ and the real number of sensors (taxels) $n_r=25$ with the following equation:
\begin{equation}
    \Omega = \frac{n_v}{n_r} = \frac{\nicefrac{A}{A_v}}{n_r},
\end{equation}
where $A=26$\,mm $\cdot 26$\,mm is the center sensing area of interest, and $A_v= \pi \cdot \sigma_{P_x} \cdot \sigma_{P_y}$ is the virtual taxel area calculated as an ellipse  with radii of the position error (standard deviations) in $x$ and $y$ directions.

\end{document}